\documentclass[lettersize,journal]{IEEEtran}

\usepackage{amsmath,amsfonts}
\usepackage{extarrows}
\usepackage{array}
\usepackage[caption=false,font=normalsize,labelfont=sf,textfont=sf]{subfig}
\usepackage{textcomp}
\usepackage{stfloats}
\usepackage{url}
\usepackage{verbatim}
\usepackage{graphicx}
\usepackage[nocompress]{cite}
\usepackage[linesnumbered,ruled,vlined]{algorithm2e}
\usepackage{booktabs}
\usepackage{xcolor}
\usepackage{hyperref}
\usepackage{marvosym}
\usepackage{footmisc}
\usepackage{xspace}
\newcommand{\ie}{{\emph{i.e.}},\xspace}
\newcommand{\eg}{{\emph{e.g.}},\xspace}
\definecolor{green}{HTML}{68BC36}

\definecolor{pur}{HTML}{9E7CC7}
\usepackage{multirow}
\usepackage[normalem]{ulem}
\usepackage{float}
\usepackage{multicol}

\usepackage{amsmath,amsfonts,amssymb,bm,mathtools}

\definecolor{DDLColor}{rgb}{1.0,0.1,0.1}

\definecolor{XZColor}{rgb}{1.0,0.0,0.0}

\def\1{\bm{1}}

\def\vepsilon{{\bm{\epsilon}}}

\def\vg{{\bm{g}}}

\def\vx{{\bm{x}}}

\DeclareMathAlphabet{\mathsfit}{\encodingdefault}{\sfdefault}{m}{sl}
\SetMathAlphabet{\mathsfit}{bold}{\encodingdefault}{\sfdefault}{bx}{n}

\newcommand{\bI}{\mathbf{I}}

\begin{document}

\title{TV-3DG: Mastering Text-to-3D Customized Generation with Visual Prompt}

\author{
Jiahui~Yang,
Donglin Di,
Baorui Ma,
Xun~Yang$^\star$,
Yongjia~Ma,
Wenzhang~Sun,
Wei~Chen, \\
Jianxun Cui$^\star$,
Zhou Xue,
Meng Wang,~\IEEEmembership{Fellow,~IEEE},
Yebin Liu,~\IEEEmembership{Member,~IEEE}

\thanks{$^\star$Corresponding authors: Jianxun Cui and Xun Yang.}
\thanks{Jiahui Yang and Jianxun Cui are with the Harbin Institute of Technology, Harbin, 150001, Heilongjiang, China.
Email: \href{mailto: yjhboyzsjdpy@gmail.com}{yjhboyzsjdpy@gmail.com}, \href{mailto: cuijianxun@hit.edu.cn}{cuijianxun@hit.edu.cn}
}

\thanks{Donglin Di, Jiahui Yang, Yongjia Ma, Wenzhang Sun and Wei Chen are with Space AI, Li Auto, 101399, Beijing, China.
Email: \href{mailto: didonglin@lixiang.com, yangjiahui1@lixiang.com, mayongjia@lixiang.com, sunwenzhang@lixiang.com, chenwei10@lixiang.com}{\{didonglin, yangjiahui1, mayongjia, sunwenzhang, chenwei10\}@lixiang.com}
}
\thanks{Baorui Ma is with School of Software, Tsinghua University, 100084, Beijing, China. Email: \href{mailto: mabaorui2014@gmail.com}{mabaorui2014@gmail.com}
}

\thanks{Xun Yang is with School of Information Science and Technology, University of Science and Technology of China, Hefei, 230026, Anhui, China. Email: \href{mailto: xyang21@ustc.edu.cn}{xyang21@ustc.edu.cn}
}

\thanks{Meng Wang is with School of Computer Science and Information Engineering, Hefei University of Technology, Hefei, 230009, Anhui, China. Email: \href{mailto: wangmeng@hfut.edu.cn}{wangmeng@hfut.edu.cn}
}

\thanks{Zhou Xue and Yebin Liu is with Department of Automation,
Tsinghua University, Beijing 100084, China. Email: \href{mailto: xuezhou08@gmail.com}{xuezhou08@gmail.com}, \href{mailto: liuyebin@mail.tsinghua.edu.cn}{liuyebin@mail.tsinghua.edu.cn}
}

}

\maketitle

\begin{abstract}

In recent years, advancements in generative models have significantly expanded the capabilities of text-to-3D generation. Many approaches rely on Score Distillation Sampling (SDS) technology. However, SDS struggles to accommodate multi-condition inputs, such as text and visual prompts, in customized generation tasks. 
To explore the core reasons, we decompose SDS into a difference term and a classifier-free guidance term. Our analysis identifies the core issue as arising from the difference term and the random noise addition during the optimization process, both contributing to deviations from the target mode during distillation.
To address this, we propose a novel algorithm, Classifier Score Matching (CSM), which removes the difference term in SDS and uses a deterministic noise addition process to reduce noise during optimization, effectively overcoming the low-quality limitations of SDS in our customized generation framework. Based on CSM, we integrate visual prompt information with an attention fusion mechanism and sampling guidance techniques, forming the Visual Prompt CSM (VPCSM) algorithm. Furthermore, we introduce a Semantic-Geometry Calibration (SGC) module to enhance quality through improved textual information integration. We present our approach as TV-3DG, with extensive experiments demonstrating its capability to achieve stable, high-quality, customized 3D generation.
Project page: \url{https://yjhboy.github.io/TV-3DG}
\end{abstract}

\section{Introduction}

\begin{figure}[t]
    \centering
    \includegraphics[width=\linewidth]{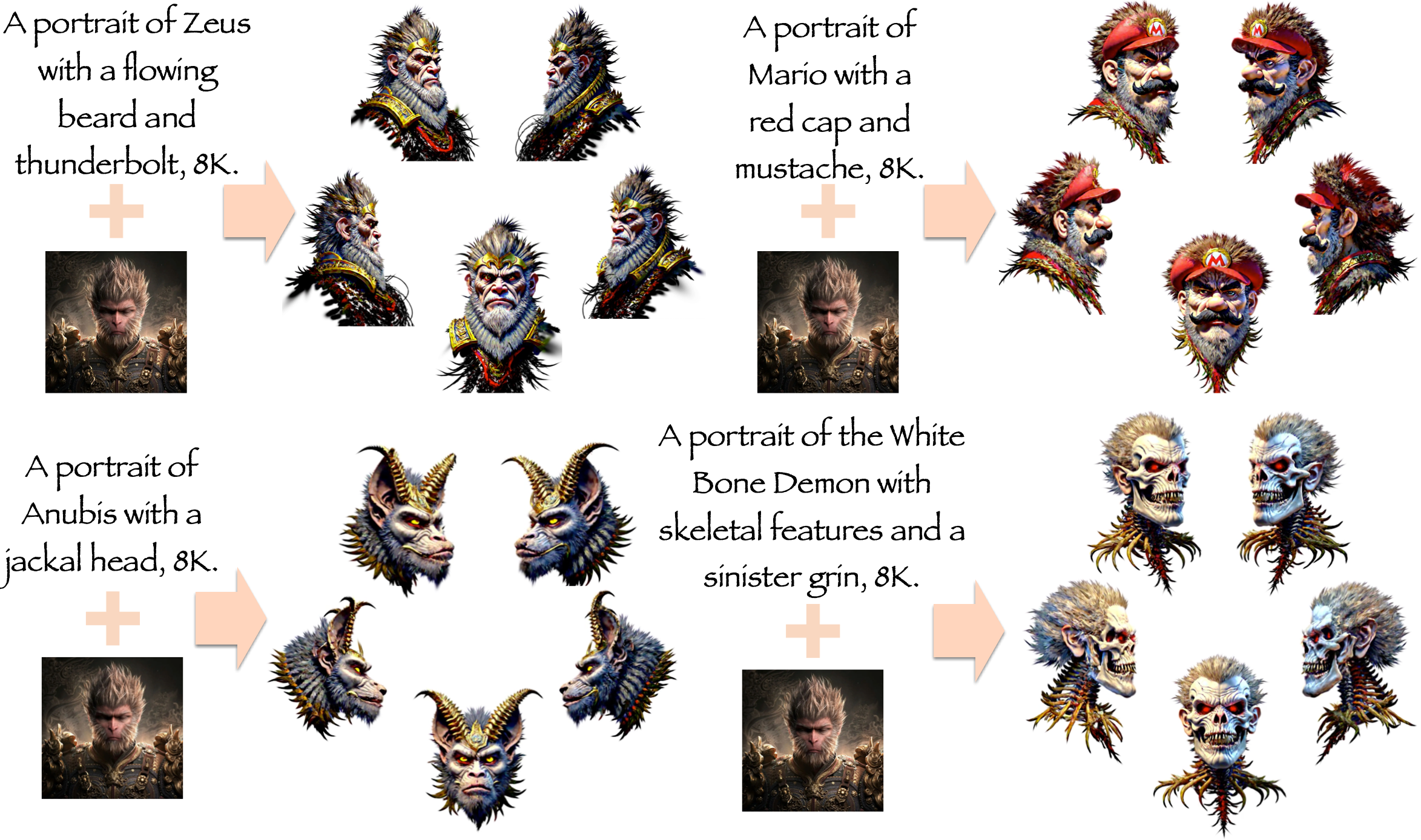}
    \caption{\textbf{An overarching understanding of our TV-3DG system.} 
    Our customized generation framework can achieves high-quality and intricate stylized generation through the use of visual prompt.}
    \label{fig:overall}
\end{figure}

\IEEEPARstart{H}{igh}-quality customized 3D content generation technology is indispensable in the digital era, characterized by extensive public participation. 
It plays pivotal roles in multimedia applications such as virtual and augmented reality, robotics, film-making, and gaming.
In these applications, users may wish to generate 3D content that not only meets vague text descriptions but also matches the style and appearance of a given visual prompt, achieving customized 3D generation, as illustrated in Fig.~\ref{fig:overall}.
While substantial attention has been devoted to controllable text-to-image (T2I) generation \cite{zhang2023adding,dhariwal2021diffusion,ho2022classifier,ye2023ip,rombach2022high,ruiz2023dreambooth}, efforts to explore high-quality customized 3D generation remain relatively under-explored. Moreover, the existing works in this domain still fall short of achieving truly high-quality customized 3D  generation.

Recently, advancements in large-scale 3D datasets \cite{chang2015shapenet,uy2019revisiting,deitke2023objaverse}, neural 3D representations \cite{mildenhall2021nerf,kerbl20233d}, and diffusion-based generative models \cite{ho2020denoising,song2020denoising,rombach2022high,karras2022elucidating} have enabled recent works \cite{nichol2022point,poole2022dreamfusion,chen2023text,lin2023magic3d,chen2023fantasia3d,tang2024lgm,wang2024prolificdreamer,liang2023luciddreamer,Tang_make_it_3d,tang2023dreamgaussian} to achieve imaginative 3D generation from text prompts. 
These methods can be categorized into optimization-based approaches \cite{poole2022dreamfusion,liang2023luciddreamer,wang2024prolificdreamer,yu2023csd,sun2023dreamcraft3d,chen2023text,tang2023dreamgaussian} and feed-forward approaches \cite{hong2023lrm,tang2024lgm,xu2024grm,liu2023zero1to3,zhang2024clay,xu2024sparp} for generating 3D results from text. A pioneering example of the optimization-based approach is Dreamfusion \cite{poole2022dreamfusion}, which introduced the Score Distillation Sampling (SDS) technique to elevate the text-to-2D priors of diffusion models into 3D generation, nearly achieving open-world 3D generation, though it suffers from issues like over-saturation, over-smoothing, and lack of detail \cite{wang2024prolificdreamer}. Conversely, feed-forward approaches, which often use multi-view prediction or sparse-view reconstruction for rapid generation, struggle with scaling model parameters due to 3D dataset limitations, resulting in lower 3D visual quality compared to optimization-based methods.
While text prompts provide some control over the generated 3D assets, producing high-fidelity and customized 3D content remains challenging due to the nature ambiguity of text.
Subsequent studies \cite{raj2023dreambooth3d,chen2023control3d,liu2024make,chen2024vp3d,zeng2023ipdreamer,chen2024generic,zhang2024clay,wang2024themestation} have explored 3D controllable generation by incorporating key text instructions or augmenting text prompts with reference images. 
Notably, VP3D \cite{chen2024vp3d} and IPDreamer \cite{zeng2023ipdreamer} achieved style-aware 3D generation by leveraging a customization model \cite{ye2023ip} and SDS \cite{poole2022dreamfusion} to optimize NeRF \cite{mildenhall2021nerf} representations. 
However, we find that these methods suffer from issues such as oversmoothing, which stem from the challenge of balancing customized style similarity with the semantic alignment of text prompts. This issue arises partly because the SDS algorithm is designed for text-to-3D generation rather than for accommodating both text and visual image conditions, leading to poor compatibility with multiple conditions. Additionally, SDS inherently exhibits problems like oversaturation and oversmoothing. We believe the core reasons for SDS's poor compatibility with multiple conditions are: (1) the presence of difference term, which exacerbate deviations from the target mode during the distillation process, and (2) the optimization mechanism that introduces random noise at time $t$, adding inherent uncertainty to the diffusion process.

In this paper, we focus on achieving high-quality customized generation through visual prompts. To address the aforementioned issues, we propose a new algorithm called Classifier Score Matching (CSM), which enables more stable and high-quality customized generation under multi-condition inputs. 
To mitigate the impact of difference term in SDS, we conduct a comprehensive analysis of SDS's optimization performance on 2D noise with both text and visual prompts (as shown in Fig.~\ref{fig:sds-analysis}), and propose removing difference term to reduce deviations from the target mode during the distillation process. 
To decrease the uncertainty introduced by random noise at time $t$, we employ a deterministic DDIM inverse noise addition. This allows CSM to better accommodate both text and visual prompt conditions.

For customized 3D generation, we utilize visual prompts for guidance. With CSM’s enhanced multi-condition compatibility, we employ an attention fusion mechanism to handle multi-condition input effectively. By integrating Classifier Free Guidance (CFG) \cite{ho2022classifier} and Perturbed Attention Guidance (PAG) \cite{ahn2024self} sampling techniques, we develop the CSM algorithm with visual prompts, referred to as VPCSM.
Additionally, we introduce a semantic-geometry calibration (SGC) module to fully leverage textual information for 3D generation guidance. 
By achieving dual alignment in semantics and geometry, we aim to further enhance the quality of 3D generation.
Finally, we name our framework TV-3DG, a novel customized generation framework designed to leverage both text descriptions and single image guidance using 3D Gaussian Splatting (3DGS) representation \cite{kerbl20233d}. As shown in Fig.~\ref{fig:overall}, our method is capable of producing high-quality and intricate stylized 3D models via a provided visual prompt.

To evaluate our approach, we conducted various qualitative and quantitative experiments in both text-to-3D and stylized 3D generation tasks. 
The results demonstrate that TV-3DG achieves superior performance in terms of fidelity and customization, validating the effectiveness of our proposed framework.
Overall, our contributions are as follows:

\begin{itemize}
\item We conduct a comprehensive analysis of SDS in customized generation and find that SDS struggles to accommodate both text and visual multi-conditions. The primary reason is the difference term and the random noise addition mechanism during the optimization process at time $t$, which together lead to deviations from the target mode during distillation.
\item To address these shortcomings in SDS, we propose Classifier Score Matching (CSM), which removes the difference term and employs a multi-step deterministic noise addition at the $t$ level, achieving a better balance between text and visual prompts in customized generation.
\item Building on CSM, we integrate visual information to develop Visual Prompt CSM (VPCSM) for customized generation. Additionally, we propose a semantic-geometry calibration (SGC) module to ensure more realistic geometry and semantics aligned with textual and visual prompts, resulting in a unified framework (TV-3DG) for high-quality customized 3D generation.
\end{itemize}

\section{Related Work}

\textbf{Text-to-3D Generation.}
One approach to text-to-3D generation involves utilizing extensive data \cite{chang2015shapenet,deitke2023objaverse} to train end-to-end 3D generative models \cite{jun2023shap,nichol2022point,zou2023triplane,hong2023lrm,tang2024lgm,xu2024grm}. However, these methods face limitations due to the scale and quality of paired text-3D data, restricting their ability to achieve specific customization. Additionally, the distribution biases in 3D datasets often lead to generated content that skews towards certain styles, such as cartoonish appearances, resulting in a lack of realism.
With advancements in 2D diffusion models \cite{ho2020denoising,rombach2022high,song2020denoising,song2020score}, researchers explore methods to transfer the strong priors of 2D models into 3D representations \cite{mildenhall2021nerf,kerbl20233d}, bypassing the need for extensive paired text-3D datasets. DreamFusion \cite{poole2022dreamfusion} leads this exploration by introducing SDS, which distills 3D assets from pre-trained 2D text-to-image diffusion models. 
Following this advance, numerous subsequent works have endeavored to further enhance text-to-3D generation.
Some studies focus on analyzing and improving SDS \cite{hong2024debiasing,wang2024prolificdreamer,liang2023luciddreamer,zhong2024dreamlcm}, while others explore different 3D representation methods to enhance visual quality \cite{chen2023text,tang2023dreamgaussian,chen2024vp3d,chen2023fantasia3d}.
For instance, Prolificdreamer \cite{wang2024prolificdreamer} aims to produce high-quality text-to-3D generation by introducing Variational Score Distillation (VSD) to address over-saturation, over-smoothing, and low-diversity issues, enhancing sample quality and diversity through a particle-based variational framework and improvements in distillation time schedule and density initialization, GSGEN \cite{chen2023text} leverages 3DGS for high-quality text-to-3D generation, addressing inaccuracies in geometry and time-consuming processes of prior methods by using a progressive optimization strategy that includes geometry optimization and appearance refinement.
Additionally, some works aim to address inconsistencies in 3D models \cite{armandpour2023re,shi2023mvdream,di2024hyper,ukarapol2024gradeadreamer}. 
MVDream \cite{shi2023mvdream} integrates the generalizability of its 2D multi-view diffusion model with the consistency of 3D renderings, enabling superior 3D generation via Score Distillation Sampling.
However, it has been empirically observed that SDS-based methods often suffer from issues such as over-saturation and over-smoothing. Additionally, relying solely on textual information is often insufficient to convey complex scene relationships or concepts, making it challenging to create customizable 3D assets that align with user expectations.
This limitation poses a potential obstacle to 3D content creation.

\textbf{Image-to-3D Generation.} Recently, numerous studies \cite{tang2023dreamgaussian,Magic123,Tang_make_it_3d,liu2023zero1to3,purushwalkam2024conrad} have explored the potential of generating 3D content from a single image. 
Magic123 \cite{Magic123} is a pioneering approach that employs a dual-prior mechanism, leveraging both 2D and 3D diffusion models \cite{rombach2022high,liu2023zero1to3}, to transform single unposed images into high-fidelity, textured 3D meshes through a coarse-to-fine optimization process.
Conrad \cite{purushwalkam2024conrad} presents a groundbreaking approach that harnesses pre-trained diffusion models to reconstruct 3D objects from a single RGB image, introducing a novel radiance field variant that explicitly captures the appearance of input image, thereby streamlining the generation of realistic 3D models.
DreamGaussian \cite{tang2023dreamgaussian} utilizes a more efficient gaussian splatting representation \cite{kerbl20233d}, greatly enhancing optimization speed. 
However, these methods produce 3D content from a single image with relatively consistent appearance, leading to a lack of flexible diversity, akin to 3D reconstruction tasks.
To mitigate this, we use a visual image to guide the text-to-3D process.

\textbf{Customized 3D Generation.}
Intuitively, customized generation can be broadly divided into content customization (object-driven) and style customization. 
In 2D generation, customized outputs are often achieved through the integration of text and image guidance using attention mechanisms \cite{hertz2024style,jeong2024visual,ye2023ip,wang2024instantstyle}. These techniques have recently been extended to 3D generation \cite{fang2024ce3d,liu2024make,chen2024vp3d,zeng2023ipdreamer,chen2024generic,liu2023stylegaussian}.
For instance, VP3D \cite{chen2024vp3d} explores the potential of stylized text-to-3D generation by leveraging visual prompts to enhance the fidelity and detail of 3D models.
Concurrently, MVEdit \cite{chen2024generic} extends 2D diffusion models for versatile and efficient 3D editing, allowing both text and image inputs to drive the generation process. 
IPDreamer \cite{zeng2023ipdreamer} focuses on appearance-controllable 3D object generation from image prompts.
Dream-in-Style \cite{kompanowski2024dream} integrates the style of a reference image into the text-to-3D generation process by manipulating features in the self-attention layers.
Additionally, ThemeStation \cite{wang2024themestation} enhances theme-aware 3D generation by synthesizing customized 3D assets based on few 3D exemplars, achieving both thematic unity and diversity through a two-stage framework and a novel dual score distillation (DSD) loss.
Make-Your-3D \cite{liu2024make} harmonizes distributions between a multi-view diffusion model and an identity-specific 2D generative model, enabling personalized, high-fidelity 3D content from a single image under different text descriptions.
While most methods primarily address object and appearance control, akin to 3D reconstruction tasks driven by text and image inputs, our work focuses on content-aware and style-aware customization, offering flexible, high-quality text-to-3D generation along with a certain degree of appearance customization. 
This approach provides a comprehensive array of options for user-oriented customized generation.

\IEEEpubidadjcol

\section{Methodology}

In this study, we present TV-3DG, an innovative framework for customized generation that utilizes text description and visual image to create high-quality, intricately 3D assets.
The structure and operational flow of the TV-3DG framework are depicted in Fig.~\ref{fig:TV-3DG}, our TV-3DG framework can be logically segmented into two modules, namely \textit{Semantic-Geometry Calibration module} (Sec.~\textcolor{red}{\ref{subsec:sgc}}), and \textit{Visual Prompt Classifier Score Matching module} (Sec.~\textcolor{red}{\ref{subsec:vpcsm}}).

\subsection{Background}\label{Background}
\textbf{Diffusion Models (DMs).}
In essence, diffusion models involve a forward/diffusion process $\{q(\vx_t|\vx_{t-1})\}_{t\in[1,T]}$ that incrementally adds noise to data points and a reverse process $\{p_{\psi}(\vx_{t-1}|\vx_t)\}_{t\in[1,T]}$ that denoises/generates the data by utilizing a predefined schedule $\beta_t$ for timestep $t$ and a learnable neural network $\psi$. 
The forward process in DMs is described as $q(\vx_t|\vx_{t-1})=\mathcal{N}(\boldsymbol{x}_t;\sqrt{1-\beta_t}\boldsymbol{x}_{t\boldsymbol{-}1},\beta_t\boldsymbol{I})$.
Given that $\bar{\alpha}_t=\prod_{i=1}^t(1-\beta_i)$, the reverse process is described as $p_\psi(\boldsymbol{x}_{t-1}|\boldsymbol{x}_t)=\mathcal{N}(\boldsymbol{x}_{t-1};\sqrt{\bar{\alpha}_{t-1}}\mu_\psi(\boldsymbol{x}_t),(1-\bar{\alpha}_{t-1})\Sigma_\psi(\boldsymbol{x}_t))$. This process begins with standard Gaussian noise $x_T := \vepsilon$ and employs a parameterized noise prediction network $\vepsilon_\psi(\vx_t, t)$ to sequentially predict the mean $\mu_\psi(\boldsymbol{x}_t)$ and variance $\Sigma_\psi(\boldsymbol{x}_t)$ of $\vx_{t-1}$ at each timestep $t$, aiming to progressively approach $\vx_0$.

DMs have demonstrated significant success in generating images from textual descriptions \cite{rombach2022high,song2020denoising,ramesh2022hierarchical}. 
In this context, the noise prediction model $\vepsilon_\psi(\vx_t, t, y)$ leverages a text prompt $(y)$ for conditioning, enhancing the generation process.
Classifier-free guidance (CFG) \cite{ho2022classifier} constitutes a pivotal technique in steering DMs towards generating outputs that adhere to specified conditions, by adjusting the predicted noise as
${\hat\vepsilon}_\psi(\vx_t,t,y)=(1+\lambda)\vepsilon_\psi(\vx_t,t,y)-\lambda\vepsilon_\psi(\vx_t,t,\emptyset)$, effectively guiding the diffusion process, where the $\emptyset$ represents empty set for the unconditional case, $\lambda>0$ is the guidance scale.

\textbf{Score Distillation Sampling (SDS).}
Given a camera parameter $c$, a differentiable renderer $\vg(\cdot,c)$ and a 3D representation with parameter $\theta$, the rendered image can be succinctly denoted as $\vx_0 := \vg(\theta,c)$. 
Then the forward process $q(\vx_t|\vx_0)$ in DMs can be recursively derived by repeatedly applying the reparameterization trick \cite{kingma2013auto}, yielding 
\begin{equation}\label{eq:ddpm-forward}
\begin{aligned}
    &q(\vx_t|\vx_0)=\mathcal{N}(\vx_t;\sqrt{\bar{\alpha}_t}\vx_0,(1-\bar{\alpha}_t)\bI) \\
    &\vx_t = \sqrt{\bar\alpha_t}\vx_0 + \sqrt{1-\bar\alpha_t}\vepsilon, \vepsilon \sim \mathcal{N}(0,\mathrm{I}).
\end{aligned}
\end{equation}
SDS is notable for its efficacy in using off-the-shelf DMs to distill 3D representations $\theta$ by minimizing a KL divergence  $\mathbb{E}_t\left[w(t)\mathrm{KL}(q(\vx_t|g(\theta);y,t)\|p_\psi(\vx_t;y,t))\right]$.
Furthermore, the loss can be expressed in the following form:
\begin{equation}\label{eq:sds-loss}
    \mathcal{L}_{\mathrm{SDS}}(\theta):=\mathbb{E}_{t,c}\left[\omega(t)||\hat\vepsilon_\psi(\boldsymbol{x}_t,t,y)-\boldsymbol{\epsilon}||_2^2\right]
\end{equation}
Ignoring the UNet Jacobian term \cite{poole2022dreamfusion}, the derivative of the SDS loss with respect to $\theta$ is computed as follows:
\begin{equation}\label{eq:sds-gradient}
\begin{split}
\nabla_\theta\mathcal{L}_{SDS}(\theta) &\approx \mathbb{E}_{t,\vepsilon,c}\left[\omega(t)(\hat\vepsilon_\psi(\vx_t,t,y) - \vepsilon)\frac{\partial \vg(\theta,c)}{\partial\theta}\right] 
\end{split}
\end{equation}
where $\omega(t)$ denotes a weighting function that is parametrized by $t$. 

Although the SDS technique facilitates open-vocabulary 3D generation, it still encounters issues such as over-smoothing and Janus problem \cite{di2024hyper,wu2024consistent3d,yu2023csd}.
\begin{figure*}[tpbh]
    \centering
    \includegraphics[width=\linewidth]{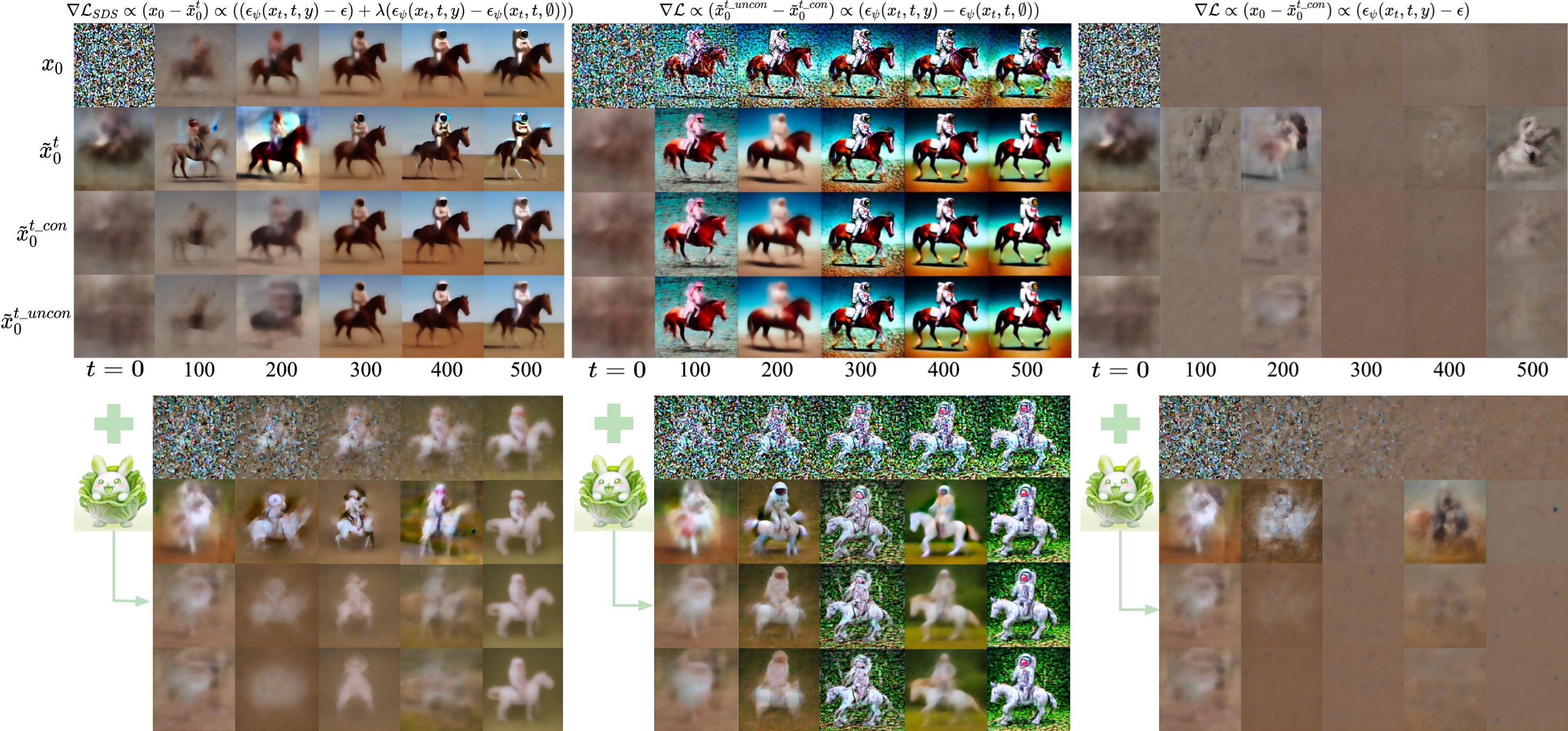}
    \caption{\textbf{In-depth analysis of SDS loss gradient in customized generation.} We use randomly initialized noise as the image. At the 2D level, we experiment with different combinations of terms in the SDS loss. 
    The left column shows results using the complete SDS loss, the middle column retains only the term with the CFG \cite{ho2022classifier} coefficient, and the right column retains only the term without the CFG coefficient, namely the difference term. We present the results guided by an arbitrary visual prompt in the lower section (method described in Sec~\ref{subsec:vpcsm}). 
    The prompt is ``A photograph of an astronaut riding a horse."
    }
    \label{fig:sds-analysis} 
\end{figure*}
\subsection{Deconstructing SDS}\label{subsec:decon-sds}
In practice, the predicted noise in SDS is subjected to CFG \cite{ho2022classifier}. The term $\hat\vepsilon_\psi(\vx_t,t,y)$ from Eq.~\ref{eq:sds-loss} can be rewritten in its form prior to applying CFG as 
$\vepsilon_\psi(\vx_t,t,y) + \lambda(\vepsilon_\psi(\vx_t,t,y) - \vepsilon_\psi(\vx_t,t, \emptyset))$.
From Eq.~\ref{eq:sds-loss}, we can infer that the objective of SDS is to minimize the discrepancy between the posterior conditional predicted noise term and the random noise term. This minimization is guided by Eq.~\ref{eq:sds-gradient} to facilitate the mode-seeking process for effective 3D distillation learning. To enhance the clarity of our analysis, we further transition Eq.~\ref{eq:sds-loss} from the noise level to the optimization objective $\vg(\theta,c)$ level.
In the context of DDIM \cite{song2020denoising}, the reverse process can be clearly understood through the application of Tweedie's formula, as follows:
\begin{equation}\label{eq:ddim-reverse}
\begin{aligned}
\boldsymbol{x}_{t-1} &= \sqrt{\bar{\alpha}_{t-1}}\tilde{\boldsymbol{x}}_0^t +
\sqrt{1-\bar{\alpha}_{t-1}-\eta^2\beta_t^2}\boldsymbol{\epsilon}_\psi(\boldsymbol{x}_t,t) + \eta\beta_t\boldsymbol{\epsilon} \\
&\xlongequal{\eta\beta_t = 0} \sqrt{\bar{\alpha}_{t-1}}\tilde{\boldsymbol{x}}_0^t + \sqrt{1-\bar{\alpha}_{t-1}}\boldsymbol{\epsilon}_\psi(\boldsymbol{x}_t,t),
\end{aligned}
\end{equation}
where 
\begin{equation}\label{eq:ddim-reverse-x0}
\tilde{\vx}_0^t=(\vx_t-\sqrt{1-\bar{\alpha}_t}\vepsilon_\psi(\vx_t,t))/\sqrt{\bar{\alpha}_t}.
\end{equation}
The notation $\vx_0^t$ represents the prediction of the target data $\vx_0$ at $t=0$ from the noise level $\vx_t$ at timestep $t$. Deterministic sampling is achieved when the term $\eta\beta_t$ is set to zero, ensuring that there is no additional stochasticity introduced during the sampling process.

Based on the formulas in Eq.~\ref{eq:ddpm-forward}, Eq.~\ref{eq:sds-gradient}, Eq.~\ref{eq:ddim-reverse}, Eq.~\ref{eq:ddim-reverse-x0} and signal-to-noise ratio (SNR) $\mathrm{SNR}(t)=\frac{\bar{\alpha}_t}{1-\bar{\alpha}_t}$, we can further expand the loss form of SDS:
\begin{equation}
\begin{split}
    &\mathcal{L}_{\mathrm{SDS}}(\theta)=\mathbb{E}_{t,c}\left[\omega(t)\|\hat\vepsilon_\psi(\boldsymbol{x}_t,t,y)-\boldsymbol{\epsilon}\|_2^2\right] \\
    &=\mathbb{E}_{t,c}[\omega(t)\sqrt{\scriptstyle \mathrm{SNR}(t)}\|\vx_{0}-\tilde{\vx}_{0}^{t\_con} +\lambda\left(\tilde{\vx}_{0}^{t\_unon} -\tilde{\vx}_{0}^{t\_con}\right)\|_{2}^{2}]
\end{split}\label{eq:sds-x0}
\end{equation}
where $\vx_{0} = \frac{\vx_{t}-\sqrt{1-\bar{\alpha}_{t}} \vepsilon }
{\sqrt{\bar\alpha_{t}}}$, $\tilde{\vx}_{0}^{t\_uncon} = \frac{\vx_{t}-\sqrt{1-\bar\alpha_{t}}\vepsilon_{t}(\vx_t,t,\emptyset)}{\sqrt{\bar\alpha_{t}}}$ and $\tilde{\vx}_{0}^{t\_con} = \frac{\vx_{t}-\sqrt{1-\bar\alpha_{t}}\vepsilon_{t}(\vx_t,t,y))}{\sqrt{\bar\alpha_{t}}}$.
Furthermore, we define $\tilde{\vx}_{0}^{t} := \tilde{\vx}_{0}^{t\_con} + \lambda (\tilde{\vx}_{0}^{t\_con} - \tilde{\vx}_{0}^{t\_uncon})$ to represent the result after applying CFG. 
Based on Eq.~\ref{eq:sds-loss}, Eq.~\ref{eq:sds-gradient} and Eq.~\ref{eq:sds-x0}, we can express the gradient of SDS as follows:
\begin{equation}\label{eq:sds-gradient-new}
\begin{aligned}
&\nabla_\theta\mathcal{L}_{SDS}(\theta) \\
&= \mathbb{E}_{t,c}\left[\omega(t)(\underbrace{\vx_{0}-\tilde{\vx}_{0}^{t\_con}}_{\delta_{dif}}+\lambda(\underbrace{\tilde{\vx}_{0}^{t\_unon} -\tilde{\vx}_{0}^{t\_con}))}_{\delta_{cfg}}\frac{\partial \vg}{\partial\theta}\right] 
\end{aligned}
\end{equation}

Based on the above analytical perspective, we visualized the changes of $\vx_{0}$, $\tilde{\vx}_{0}^{t\_uncon}$, $\tilde{\vx}_{0}^{t\_con}$, and $\tilde{\vx}_{0}^t$ during the SDS optimization process, as shown in Fig.~\ref{fig:sds-analysis}. From Eq.~\ref{eq:sds-x0}, we know that the optimization goal of SDS is to optimize the 3D parameter $\theta$ to minimize the difference between $\vg(\theta, c)$ and $\tilde{\vx}_0^t$. From the results in Fig.~\ref{fig:sds-analysis}, we can see that the states of $\tilde{\vx}_{0}^{t\_uncon}$ (fourth row) and $\tilde{\vx}_{0}^{t\_con}$ (third row) are almost identical during the optimization process, whereas there are differences between $\vx_0$ and $\tilde{\vx}_0^t$. Therefore, the mismatch between $\vx_0$ and $\tilde{\vx}_0^t$ mainly comes from the term $(\vx_{0} - \tilde{\vx}_{0}^{t\_con})$. For convenience, we define $\delta_{dif} := \vx_{0} - \tilde{\vx}_{0}^{t\_con}$ to represent the difference term, and $\delta_{cfg} := \tilde{\vx}_{0}^{t\_uncon} - \tilde{\vx}_{0}^{t\_con}$ to represent the term with the CFG coefficient. 
To investigate the impact of different terms, we present the 2D optimization process with only text condition and with the addition of a visual prompt in the top and bottom layouts of Fig.~\ref{fig:sds-analysis}, respectively. The left side displays the complete SDS loss terms, while the middle and right sides show the results using only $\delta_{cfg}$ and $\delta_{dif}$. It can be observed that the core contributing factor is the $\delta_{cfg}$ loss term, while the $\delta_{dif}$ term, which contributes minimally to the overall optimization, is the main factor causing the noise term mismatch in Eq.~\ref{eq:sds-loss}.
Eliminating the difference term $\delta_{dif}$ reduces inaccuracies in the distillation process, as demonstrated in Fig.~\ref{fig:sds-analysis} at $t=500$ on the left and middle. The results in the upper section of Fig.~\ref{fig:sds-analysis}, which include only the $\delta_{cfg}$ term, more closely align with the text description. Meanwhile, the lower section more clearly reflects the visual prompt information while preserving the text semantics.
Therefore, we propose using $\delta_{cfg}$ to ensure a better optimization process.
Nonetheless, the results presented in Fig.~\ref{fig:sds-analysis} remains quite vague and lacks detail, which is still unacceptable.

\begin{figure}[t]
    \centering
    \includegraphics[width=\linewidth]{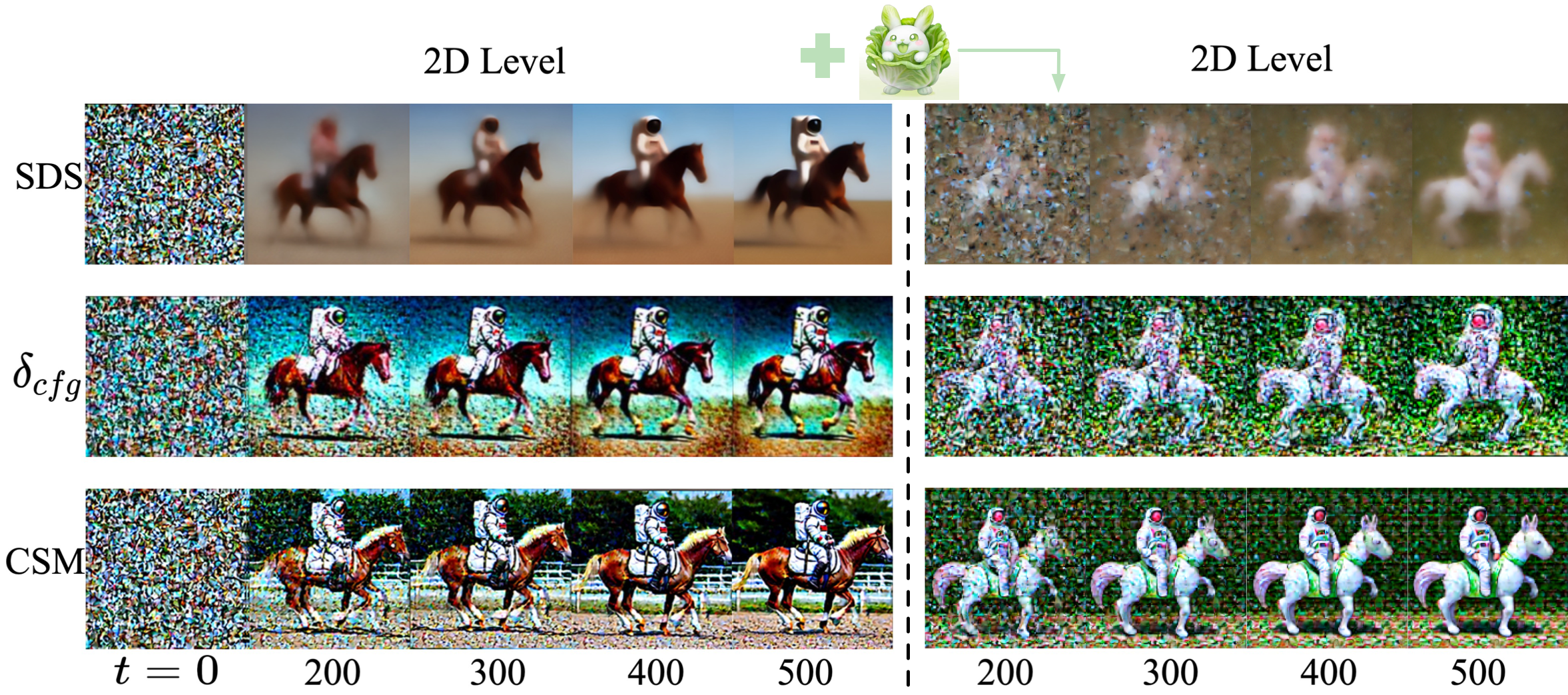}
    \caption{\textbf{Evaluation of Classifier Score Matching (CSM) loss.} We conduct experiments with our CSM loss on 2D level. We present the optimization process of $\vx_0$ and observe that CSM achieves clearer image details compared to SDS loss at the same timestep. When a visual prompt with significantly different semantics is introduced, CSM effectively preserves clear geometric structures and captures enhanced style and texture information.
    }
    \label{fig:csm-sds} 
\end{figure}

\subsection{Classifier Score Matching (CSM)}\label{subsec:csm}
From the expressions of $\tilde{\vx}_{0}^{t\_con}$ and $\tilde{\vx}_{0}^{t\_uncon}$, it is evident that both are predicted based on the noise level $\vx_t$ at time $t$. Therefore, the quality of the loss gradient is primarily influenced by adjustments at the $\vx_t$ level.
We propose that using the deterministic noise addition method of DDIM \cite{song2020denoising} inversion, instead of the stochastic noise addition method in SDS's DDPM \cite{ho2020denoising}, will yield better results at the $\vx_t$ level. In DDPM, noise is randomly sampled from a normal distribution at each step of the noise addition process. In contrast, DDIM inversion leverages the model's generative capabilities more effectively during the noise addition process, ensuring a more accurate and matched correspondence between each intermediate result $\vx_t$ and the initial image $\vx_0$. This precise correspondence allows the model to predict and remove noise more accurately during the reverse diffusion process, thereby generating higher quality images.
This assertion is supported by the comparative experiments on the 2D level shown in Fig.~\ref{fig:csm-sds}. In particular, given a timestep interval $\delta t$, DDIM inversion adds noise to $\vx_0$ to obtain $\vx_t$ through the following formula:
\begin{equation}
\begin{aligned}
&\tilde\vx_0^{t-\delta t} = \frac{\vx_{t-\delta t}-\sqrt{1-\bar\alpha_{t-\delta t}}\epsilon_\psi(\vx_{t-\delta t},t-\delta t,\emptyset)}{\sqrt{\bar\alpha_{t-\delta t}}}\\
&\boldsymbol{x}_t=\sqrt{\bar{\alpha}_t}\tilde{\boldsymbol{x}}_0^{t-\delta t}+\sqrt{1-\bar{\alpha}_t}\boldsymbol{\epsilon}_\psi(\boldsymbol{x}_{t-\delta t},t-\delta t,\emptyset) 
\end{aligned}
\end{equation}

\begin{equation}\label{eq:csm}
\begin{aligned}
\nabla_{\theta}\mathcal{L}_{CSM}=\mathbb{E}_{t,c}[\omega(t)\lambda(\vepsilon_{\psi}(\vx_{t}^{inv},t,y)-\vepsilon_{\psi}(\vx_{t}^{inv},t,\emptyset))\frac{\partial \vg}{\partial\theta}]
\end{aligned}
\end{equation}

\begin{figure}[t]
    \centering
    \includegraphics[width=\linewidth]{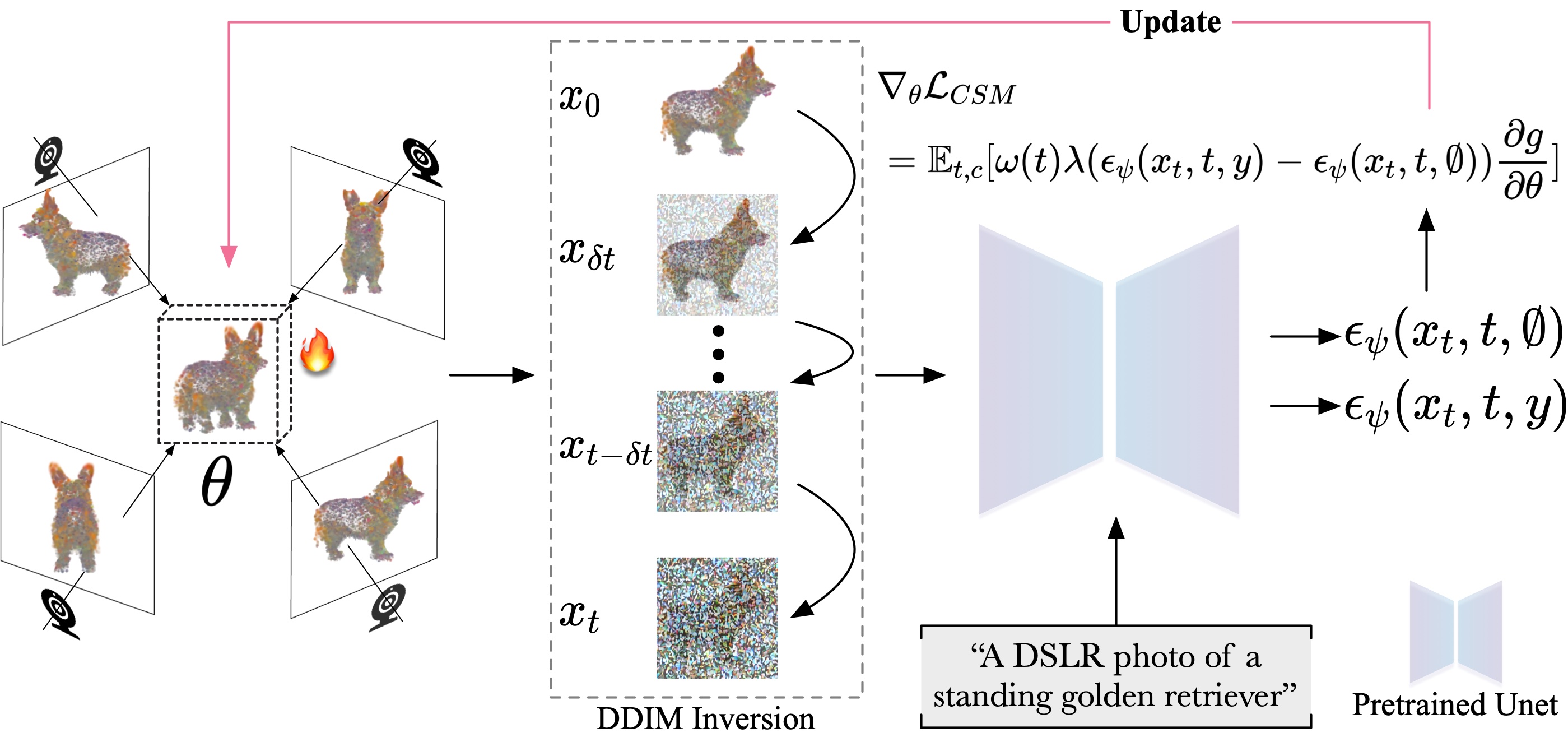}
    \caption{\textbf{Illustration of Classifier Score Matching (CSM).} We aim to utilize a pre-trained text-to-image model $\vepsilon_\psi$ to perform score matching on the 2D level. An image is rendered from $\theta$ for a specific viewpoint, which is then subjected to noise addition through DDIM inversion. The denoising Unet subsequently estimates the noise. In our framework, it is necessary to estimate two outputs of the Unet: $\vepsilon_\psi(\vx_t,t,y)$ and $\vepsilon_\psi(\vx_t,t,\emptyset)$, with a classifier-free guidance scale $\lambda$. Finally, optimization is performed using our proposed CSM.}
    \label{fig:csm} 
\end{figure}

We denote the result of DDIM inversion $\vx_t$ as $\vx_t^{inv}$. The denoising Unet then calculates the predicted noise $\vepsilon_{\psi}(\vx_{t}^{inv}, t, y)$ and $\vepsilon_{\psi}(\vx_{t}^{inv}, t, \emptyset)$, which subsequently form the final loss, as shown in Eq.~\ref{eq:csm}. Given our approach of utilizing only the $\delta_{cfg}$ term and achieving more precise matching at the $\vx_t$ level via DDIM inversion, we designate this method as Classifier Score Matching (CSM). To validate the effectiveness of CSM, we present a toy example where, for simplicity, we set the parameter $\delta t$ to 100. We then compare the results with those obtained using SDS and the $\delta_{cfg}$ term in SDS, as shown in Fig.~\ref{fig:csm-sds}. The results indicate that CSM achieves higher quality generation on 2D level. The core computational process is illustrated in Fig.~\ref{fig:csm} and Algorithm~\ref{alg:csm}.

\begin{algorithm}[t]
\SetKwInOut{To}{to}
\SetKwInOut{Input}{Input}
\SetKwInOut{Output}{Output}
\SetKwInOut{Initialization}{Initialization}
\DontPrintSemicolon
\caption{Classifier Score Matching}
\label{alg:csm}
\Input{
Text prompt $y$; CFG scale $\lambda$; Training steps $iterations$;
Step interval $\delta t$ for DDIM inversion; Camera poses $c$ 
}
\Output{
Optimized 3DGS Model $\theta$
}

\BlankLine
\For{$ i \gets 1$ \KwTo $iterations$}
{
Differentiable rendering: $\vx_0 \gets \vg(\theta,c)$

Sample: $t\sim \mathcal{U}(0.02,0.98)$

Residual steps $\delta t_r \gets t \mod \delta t$

\eIf{$\delta t_r == 0$}{
$k \gets \frac{t}{\delta t}$;
$\delta t_r \gets \delta t$
}{
$k \gets \lfloor \frac{t}{\delta t} \rfloor + 1$
}

\For{$i \gets 1$ \KwTo $k$ }{
\eIf{$i==1$}{
$\delta t^\prime \gets \delta t_r$; $\delta t^{\prime\prime} \gets \delta t^\prime + \delta t$
}{
$\delta t^\prime \gets (i-1) \times \delta t^\prime$; $\delta t^{\prime\prime} \gets \delta t^\prime + \delta t_r$
}
$\tilde{\vx}_0^{\delta t^\prime}=\frac{\vx_{\delta t^\prime}-\sqrt{1-\bar{\alpha}_{\delta t^\prime}}\epsilon_\psi(\vx_{\delta t^\prime},\delta t^\prime,\emptyset)}{\sqrt{\bar{\alpha}_{\delta t^\prime}}}$

$\vx_{\delta t^{\prime\prime}}=\sqrt{\bar{\alpha}_{\delta t^{\prime\prime}}}\tilde{\vx}_0^{\delta t^\prime}+\sqrt{1-\bar{\alpha}_{\delta t^{\prime\prime}}}\vepsilon_\psi(\vx_{\delta t^\prime},\delta t^\prime,\emptyset)$

}

$\vx_t^{inv} \gets \vx_t$ \tcp{$t==\delta t^{\prime\prime}$}

Unet output: $\vepsilon_\psi(\vx_t^{inv},t,y)$ and $\vepsilon_\psi(\vx_t^{inv},t,\emptyset)$

$\nabla_\theta\mathcal{L}_{CSM} \propto \omega(t)\lambda (\vepsilon_\psi(\vx_t^{inv},t,y) - \vepsilon_\psi(\vx_t^{inv},t,\emptyset))$

Update $\theta \gets \nabla_\theta\mathcal{L}_{CSM}$
}
\end{algorithm}

\begin{figure*}[tpbh]
    \centering
    \includegraphics[width=\linewidth]{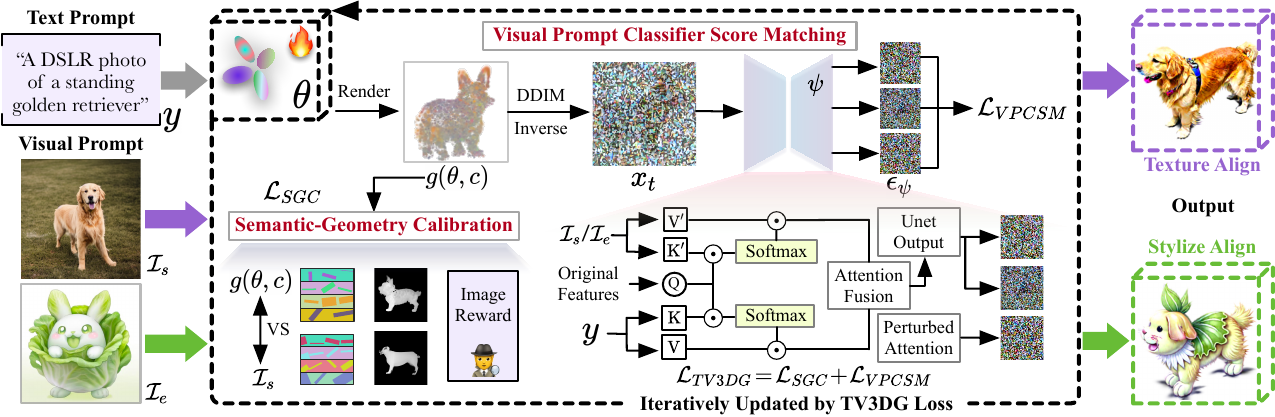}
    \caption{\textbf{Overview of our proposed TV-3DG.} 
    Our framework integrates several advanced modules: a Visual Prompt Classifier Score Matching (VPCSM) module that incorporates visual prompt guidance along with Classifier Free Guidance and Perturbed Attention Guidance techniques for aligning texture and style; and a Semantic-Geometry Calibration (SGC) module designed to enhance semantic and geometric fidelity. Our input includes a text prompt and a visual prompt. When the visual prompt aligns with the textual description, the framework generates high-quality optimized outputs (\ie texture alignment), as indicated by the \textcolor{pur}{\textbf{purple}} arrow. Conversely, when the visual prompt and text description are inconsistent, TV3DG learns relevant stylistic and appearance elements from the visual information (\ie style alignment) while retaining the main subject depicted in the text, as indicated by the \textcolor{green}{\textbf{green}} arrow.
    }
    \label{fig:TV-3DG} 
\end{figure*}

\subsection{Semantic-Geometry Calibration (SGC)}\label{subsec:sgc} 
This section aims to fully explore the potential of text prompts in customized 3D generation. Our goal is to ensure that the generated 3D content not only aligns more closely with the semantic information in the text but also matches the latent visual information in terms of geometric structure. By achieving dual alignment in both semantics and geometry, we aim to further enhance the quality of 3D generation.

\textbf{Human Feedback Image Reward Guidance.}
In the domain of text-to-image generation, extensive research has been conducted on both auto-regressive models and diffusion-based models \cite{hoogeboom2021autoregressive,ramesh2022hierarchical}. 
These approaches have demonstrated significant advancements in the fidelity and versatility of generated images.
However, a key challenge persists: the discrepancy between the noisy distributions utilized in model pre-training and the diverse distributions encountered in actual user prompts. 
This divergence hinder the ability of the model to accurately align with human preferences, resulting in a suboptimal representation of user intent in the generated imagery.
This challenge can be alleviated through the use of ImageReward \cite{xu2024imagereward}, which integrates human feedback into text-to-image models to enhance the alignment between generated images and textual descriptions.
With this in mind, we harness the ImageReward model $IR$ to guide the optimization of our text-to-3D method through human scoring.
Concretely, given a differentiable reward-to-loss module $\phi$, the loss function derived from human feedback on image rewards can be formulated as:
\begin{equation}\label{eq:ir}
    \mathcal{L}_{IR} = \mathbb{E}_{c} [\phi(IR(\mathcal{S},y))], \mathcal{S}=\begin{pmatrix}\vg(\theta,c_f)&\vg(\theta,c_r)\\\vg(\theta,c_l)&\vg(\theta,c_b)\end{pmatrix},
\end{equation}
where $\mathcal{S}$ represents a $2\times2$ grid of rendered images derived from $\theta$ under the camera conditions $\mathcal{C}=\{c_f, c_r, c_b, c_l\}$, which respectively represent the front, right, back, and left directions. 
This grid encapsulates scene information at a visual level. 
The symbol $\mathbb{E}_{c}$ signifies the mean computed over the conditions in $\mathcal{C}$.

\textbf{Visual Reconstruction Guidance.}
Prior works \cite{chen2024vp3d,yang2022banmo,yu2024boostdream,tang2023dreamgaussian,sun2023dreamcraft3d} have shown visual consistency reward is beneficial to shape appearance details. Therefore, we not only use prompt cue via ImageReward but also leverage latent visual cue corresponding to the text to generate 3D models that closely match the text description. 

Leveraging the rendered information $\mathcal{S}$, we employ multi-view images $\mathcal{M}$ obtained by processing $y$ through a multi-view generator \cite{wang2023imagedream}, which serve to guide $\mathcal{S}$ at both the semantic and geometric feature levels.
For semantic enhancement, we employ the pre-trained self-supervised vision transformer model DINO-ViT \cite{caron2021emerging} as our semantic feature extractor, denoted as $\mathcal{E}(\cdot)$. Subsequently, we apply the following formula to quantify the semantic discrepancy between the rendered image and the latent visual representation.
\begin{equation}
\mathcal{L}_{semantic} = || \mathcal{E}(\mathcal{M}) - \mathcal{E}(\mathcal{S}) ||^2
\end{equation}
For geometric enhancement, we penalize the geometric-level discrepancies, specifically in depth and normal, by employing the subsequent formula:
\begin{equation}
\begin{split}
&\mathcal{L}_{\mathrm{depth}}=-\frac{\boldsymbol{conv}(\varpi (\mathcal{M}),\varpi (\mathcal{S}))}{\boldsymbol{var}(\varpi (\mathcal{M})) \cdot \boldsymbol{var}(\varpi (\mathcal{S}))} \\
&\mathcal{L}_\mathrm{normal}=-\frac{\boldsymbol{nor}(\mathcal{M})\cdot{\boldsymbol{nor}}(\mathcal{S})}{\left\|\boldsymbol{nor}(\mathcal{M})\right\|_2\cdot\left\|{\boldsymbol{nor}}(\mathcal{S})\right\|_2},
\end{split}
\end{equation}
where $\varpi(\cdot)$ and $\boldsymbol{nor}(\cdot)$ denote the depth and normal extractors, respectively, as referenced in \cite{eftekhar2021omnidata}. The operators $\boldsymbol{var}(\cdot)$ and $\boldsymbol{conv}(\cdot)$ represent the variance and covariance, respectively. The depth loss, expressed as $\mathcal{L}_{depth}$, is formulated using a negative Pearson correlation to account for scale mismatch in depth measurements.
To streamline the terminology, we denote $\mathcal{L}_{IR}$, $\mathcal{L}_{semantic}$, $\mathcal{L}_{depth}$ and $\mathcal{L}_{normal}$ collectively as the $\mathcal{L}_{SGC}$ loss, as the following, $\lambda_1, \lambda_2, \lambda_i$ are hyperparameters.
\begin{equation}\label{eq:sgc}
\mathcal{L}_{SGC} = \lambda_1 (\mathcal{L}_{depth}+\mathcal{L}_{normal}) + \lambda_2 \mathcal{L}_{semantic} + \lambda_i \mathcal{L}_{IR},
\end{equation}
By enforcing the aforementioned guidance, our TV-3DG framework enhances the fidelity and detail of the generated 3D representation.

Hence, in the SGC phase, we employ both the reconstruction loss and the ImageReward loss to direct the update trajectory of $\theta$.

\subsection{Customized Generation via Visual Prompt}\label{subsec:vpcsm}
It is worth recalling that our goal is to achieve customized generation using visual prompts. Specifically, we aim for the generated results to maintain consistency with the text prompt in terms of core content and with the visual prompt in terms of visual appearance. Therefore, we employ the visual prompt as a conditioning input. Following the KL loss optimization objective of SDS, our customized generation framework using visual prompts aims to mathematically optimize the following KL loss:
\begin{equation}
\min_{\theta\in\Theta}\mathcal{L}(\theta) = \mathbb{E}_t\left[w(t)\mathrm{KL}(q(\boldsymbol{x}_t|g(\theta);y,t)\|p_\psi(\boldsymbol{x}_t;y,t,v))\right],
\end{equation}
where $p_\psi(\boldsymbol{x}_t; y, t, v)$ represents the reverse process conditioned on both the text prompt $y$ and the visual prompt $v$.

\textbf{Conditioning with Visual Prompts.}
To incorporate the visual prompt into the previously developed text-to-3D CSM algorithm, we employ an attention fusion mechanism \cite{hertz2022prompt,ye2023ip,jeong2024visual,hertz2024style} to integrate visual image information with textual information.
To fully utilize the visual prompt, we aim to enhance the CSM algorithm with detailed guidance for higher-quality text-to-3D generation when the visual prompt aligns with the text description. Conversely, when the visual prompt differs from the text, we seek to infuse the CSM algorithm with stylistic appearance information to achieve high-quality stylized 3D generation. These two aspects together form our customized generation framework.
In particular, to boost the generation of a 3D model, we initially employ a visual guidance image $\mathcal{I}_g$, which derived from either a self-guidance image $\mathcal{I}_s$ or a reference image $\mathcal{I}_e$, as shown in the switch component of Fig.~\ref{fig:TV-3DG}.
The self-guidance image $\mathcal{I}_s$ represents a visual prompt consistent with the textual information. It can be generated by a T2I model \cite{rombach2022high} to work together with the textual information, guiding the generation results towards higher quality.
The reference image $\mathcal{I}_e$ is an arbitrary image that a user can freely provide to influence the style of the generated output.
Then $\mathcal{I}_g$ is projected into latent space $v$ through the CLIP image encoder \cite{radford2021learning} and a multilayer perceptron \cite{ye2023ip}.
Subsequently, the attention fusion mechanism is employed to integrate the visual prompt. Within the SD framework, text features are inserted into the UNet model through cross-attention layers via the CLIP text encoder.
The attention fusion mechanism combines multi-conditional information by balancing it through the attention layers, enabling multi-condition control. Technically, the output of the attention fusion is $\mathbf{F}$:
\begin{equation}
\mathbf{F}=\mathrm{Softmax}(\frac{\mathbf{QK}^{\top}}{\sqrt{d}})\mathbf{V}+  \tau * \mathrm{Softmax}(\frac{\mathbf{Q}(\mathbf{K}^{\prime})^{\top}}{\sqrt{d}})\mathbf{V}^{\prime},
\end{equation}
where $\tau \geq 0$ is the scale of image condition, $\mathbf{Q}$ is the query matrix from query features, $\mathbf{K^\prime, V^\prime}$ is the key, value matrices from visual condition $v$, and $\mathbf{K, V}$ is the key, value matrices from text condition $y$.

The quality of condition-controlled generation is not solely dictated by the integration of multimodal inputs but is also profoundly impacted by sampling guidance techniques. Beyond leveraging Classifier-Free Guidance (CFG), we incorporate Perturbed-Attention Guidance (PAG) \cite{ahn2024self} to direct and refine the sampling process. PAG facilitates high-quality guidance without necessitating additional training by perturbing the self-attention map ($\mathbf{A}$) to an identity matrix, thereby preserving only the ($\mathbf{V}$) matrix that encapsulates appearance information \cite{balaji2022ediff,hertz2022prompt,tewel2023key}. The corresponding formulation is presented as follows:
\begin{equation}SA(\mathbf{Q},\mathbf{K},\mathbf{V})=\mathbf{AV} \longmapsto PSA(\mathbf{Q},\mathbf{K},\mathbf{V})=\mathbf{IV},\end{equation}
where $SA$ denotes self-attention and $PSA$ denotes the perturbed self-attention operation.
Subsequently, sampling guidance is performed in a manner akin to CFG, expressed as:
\begin{equation}\label{eq:pag}
    \hat{\vepsilon}(\vx_t,t,y,v)=\vepsilon(\vx_t,t,y,v)+s(\vepsilon(\vx_t,t,y,v)-\bar\vepsilon(\vx_t,t,y,v)),
\end{equation}
where $s$ is the guidance coefficient, and $\bar{\vepsilon}(\cdot)$ represents the output of the unet after applying the PSA mechanism.

\begin{figure*}[]
    \centering
    \includegraphics[width=\linewidth]{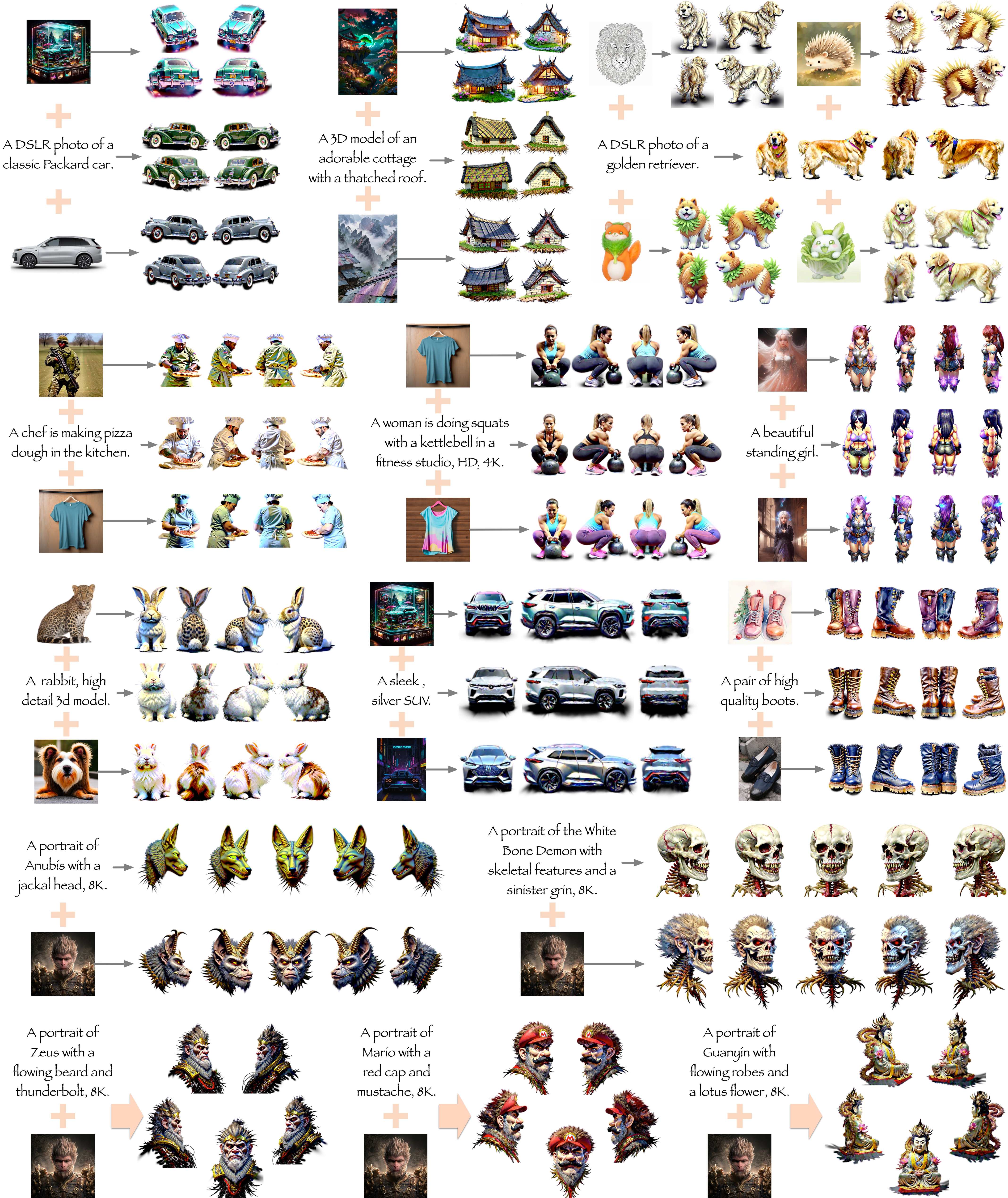}
    \caption{\textbf{Visual results of TV-3DG} with various customized text and reference visual prompts. Our method demonstrates a strong ability to generate high-quality, consistent, intricate, and style-controllable 3D assets.
    Please zoom in to view details.}
    \label{fig:visualization} 
\end{figure*}

\textbf{Visual Prompt Classifier Score Matching (VPCSM).}
Next, we use the rendered images $\mathcal{S}$ of 3DGS as the initial state $\mathcal{X}_0$ for the forward diffusion process.
Following DDIM inversion with a timestep interval $\delta t$, we obtain intermediate state $\mathcal{X}_t$.
Considering this operation as a parallel computation with a batch size of 4, we demonstrate the computation flow at the single sample level $\vx_0$ (corresponding to $\mathcal{X}_0$), thus, $\vx_0 \in \{\vg(\theta,c_i)|i \in \{f,l,r,b\}\}$.
Notably, for the timestep $t$, we employ an anneal timestep strategy. Specifically, we define a warmup phase during which the range of timesteps is linearly decreased from $[t_{min}^{up}, t_{max}^{up}]$ to $[t_{min}^{low}, t_{max}^{low}]$. During this phase, both the upper and lower limits of the timesteps decreases with the number of warmup steps, denoted as $W$, resulting in a deformable sliding window for timestep ranges. This phase is designed to focus on the construction of the global structure. After the warmup phase, timesteps are selected solely from $\left[t_{min}^{low}, t_{max}^{low}\right]$, aiming to optimize the appearance details.
Subsequently, following the procedure of CSM, we propose a text-to-3D generation framework with visual control, termed Visual Prompt Classifier Score Matching (VPCSM). VPCSM employs both CFG and PAG sampling guidance techniques and incorporates conditions from both textual and visual information. The final gradient of our introduced VPCSM loss can be derived from Eq.~\ref{eq:csm} and Eq.~\ref{eq:pag}, and is formulated as follows:
\begin{equation}\label{eq:vpcsm}
\begin{aligned}
&\nabla_\theta\mathcal{L}_{VPCSM}(\theta)\\
&=\mathbb{E}_{t,c}[\omega(t)(\lambda(\vepsilon_\psi(\vx_t^{inv},t,y,v)-\vepsilon_\psi(\vx_{t}^{inv},t,\emptyset,\emptyset)) \\
&+s(\vepsilon_\theta(\vx_t^{inv},t,y,v)-\hat{\vepsilon}_\theta(\vx_t^{inv},t,y,v)))\frac{\partial \vg(\theta,c)}{\partial\theta}],
\end{aligned}
\end{equation}
In conclusion, our customized generative framework, termed \textbf{TV-3DG}, encompasses several key components: the CSM algorithm (Eq.~\ref{eq:csm}), the enhanced alignment of geometry and semantics (Eq.~\ref{eq:sgc} and Eq.~\ref{eq:ir}), and the VPCSM framework for efficient integration of visual prompt information (Eq.~\ref{eq:vpcsm}). The framework is defined as follows:
\begin{equation}
    \mathcal{L}_{TV3DG} = \mathcal{L}_{SGC} + \mathcal{L}_{VPCSM},
\end{equation}
In summary, we can flexibly update the 3DGS parameter $\theta$ using the hyperparameters $\delta t, \tau, \lambda, s, \lambda_1, \lambda_2, \lambda_i, W$, allowing for customized guidance in style or content.

\section{Experiments}

\subsection{Experiment Setup}
\textbf{Implementation Details.}
We implement our framework based on LucidDreamer \cite{liang2023luciddreamer}.
All experiments were conducted on an A100 GPU with 80GB of VRAM.
The generation quality improves with more iterations and we find 4,000 iterations (1.1 hours on an A100 GPU, with VRAM usage approximately 22 to 25GB) already produces high-quality 3D models.
By default, we set the hyperparameters as follows: $\tau = 0.5$, $\lambda = 7.5$, $s = 1$, $\lambda_1 = 1$, $\lambda_2 = 4$, and $\lambda_i = 2.5$. The warmup steps $W$, which are utilized for the purpose of timestep annealing, are set to $W_{{1}/{3}}$, representing ${1}/{3}$ of the total iterations.
Additionally, we use the Stable Diffusion v1.5 model \cite{rombach2022high} as our pretrained text-to-image diffusion model.
The strategy for selecting timesteps involves setting both upper and lower limits. Specifically, the upper and lower bounds for the maximum timestep are set to $t_{max}^{up} = 0.98$ and $t_{max}^{low} = 0.78$ respectively. Similarly, the upper and lower bounds for the minimum timestep are set to $t_{min}^{up} = 0.22$ and $t_{min}^{low} = 0.02$. This configuration ensures a controlled and gradual reduction of the timestep range during the warmup phase, facilitating a smooth annealing process that aligns with our optimization objectives.
Additionally, within the CSM algorithm, we set $\delta t = 50$. This parameter plays a crucial role in defining the granularity of the timestep adjustments, enabling precise control over the evolution of the model dynamics.

For the text-to-3D generation task, we switch the component in Fig.~\ref{fig:TV-3DG} to connect to the T2I model, enhancing the quality of the content output with self-guidance image (\ie $\mathcal{I}_e$). Additionally, beyond generating from the T2I model, users can also opt to incorporate custom images that align with the text description for further enhancement.
Additionally, beyond generating from the T2I model, users can also opt to incorporate custom images that align with the text description for further enhancement.
For stylized generation tasks, we switch it to the user's provided reference image (\ie $\mathcal{I}_e$).

\begin{figure*}[tpbh]
    \centering
    \includegraphics[width=\linewidth]{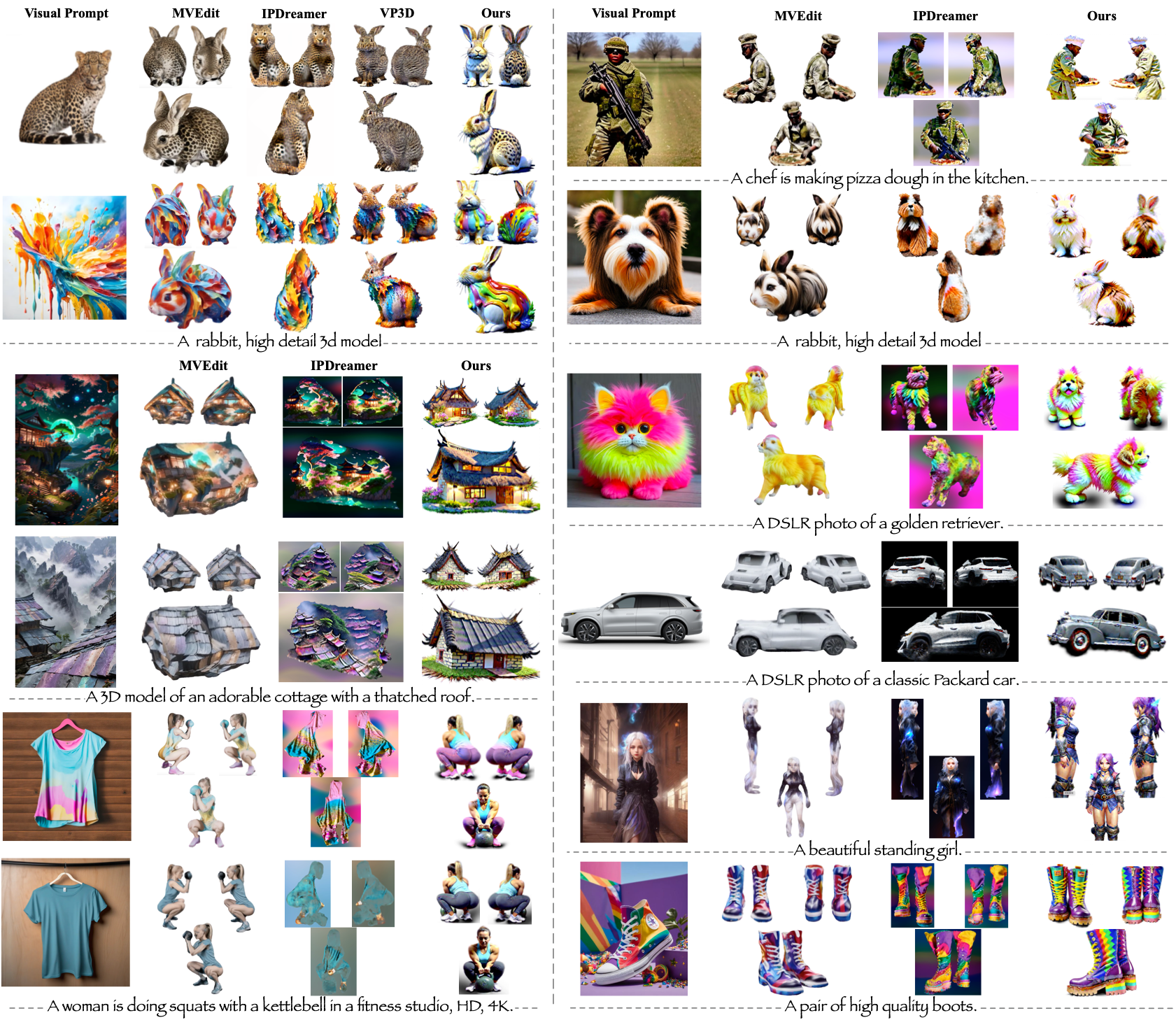}
    \caption{\textbf{Comparative analysis of 3D stylized generation task between our method and established baselines.} Experimental outcomes indicate that our approach proficiently produces stylized 3D assets. For the VP3D baseline \cite{chen2024vp3d}, since it is not open-sourced, we compare our results based on their official demo. This corresponds to the example in the top-left corner: \textit{"A rabbit, high detailed 3D model"}. Please zoom in to view details.}
    \label{fig:Comparison_ST} 
\end{figure*}

\textbf{Compared Methods.}
Given that our customized generation framework supports both text-to-3D and stylized generation, we conduct evaluations on two distinct tasks.
For the text-to-3D quality assessment task, we compare our method with text-to-3D baselines: DreamFusion \cite{poole2022dreamfusion}, Magic3D \cite{lin2023magic3d}, Fantasia3D \cite{chen2023fantasia3d}, ProlificDreamer \cite{wang2024prolificdreamer}, LucidDreamer \cite{liang2023luciddreamer} and VP3D \cite{chen2024vp3d}. 
For the stylized generation task, our comparisons include IPDreamer \cite{zeng2023ipdreamer}, MVEdit \cite{chen2024generic}, and VP3D \cite{chen2024vp3d}, where IPDreamer and VP3D enable controlled 3D object generation with image prompts, and MVEdit generats 3D object by introducing a training-free 3D Adapter that seamlessly integrates multi-view editing for controllable 3D synthesis from 2D diffusion models.
Notably, when conducting the VP3D evaluation, we used cases from its official website for a fair comparison, as it is not open-sourced, as shown in Fig.~\ref{fig:Comparison_ST}.
In addition to the aforementioned comparative methods, we also conducted a comparative analysis within our framework of a series of optimization-based methods, namely SDS \cite{poole2022dreamfusion}, CSD \cite{yu2023csd}, ISM \cite{liang2023luciddreamer}, VSD \cite{wang2024prolificdreamer}, and our own CSM approach.

\textbf{Metrics.} 
We employ the CLIP-Score metric \cite{jain2022zero} to evaluate the semantic alignment between generated images and their textual descriptions. This metric utilizes the CLIP model's joint embedding space to quantify coherence, ensuring that generated images are not only descriptively accurate but also semantically consistent.
Our methodology incorporates multiple CLIP retrieval models—ViT-B/16, ViT-B/32, and ViT-L/14 \cite{dosovitskiy2020image}—for a comprehensive evaluation. This approach ensures that our assessment is both robust and unbiased, capturing a wide range of semantic relationships.
To assess 3D consistency, we use the A-LPIPS metric \cite{zhang2018lpips}, which measures the perceptual similarity between adjacent rendered views of 3D models. By averaging these scores, we obtain a reliable measure of 3D visual coherence. We perform a detailed evaluation using the VGG \cite{simonyan2014vgg} and AlexNet \cite{krizhevsky2012alex} architectures to benchmark the sensitivity of different models in detecting perceptual inconsistencies.
We also incorporate the Fréchet Inception Distance (FID) \cite{heusel2017gans-fid} to evaluate the divergence between rendered 3D images and their corresponding 2D images derived from text. The FID score offers a statistical measure of the distance between the feature distributions of these image sets, providing insight into the visual diversity and realism of our generated images.
Additionally, we perform a user study to validate the effectiveness of our approach in terms of content fidelity, prompt adherence, style fusion effectiveness, 3D consistency, and overall quality.
\begin{figure*}[tpbh]
    \centering
    \includegraphics[width=\linewidth]{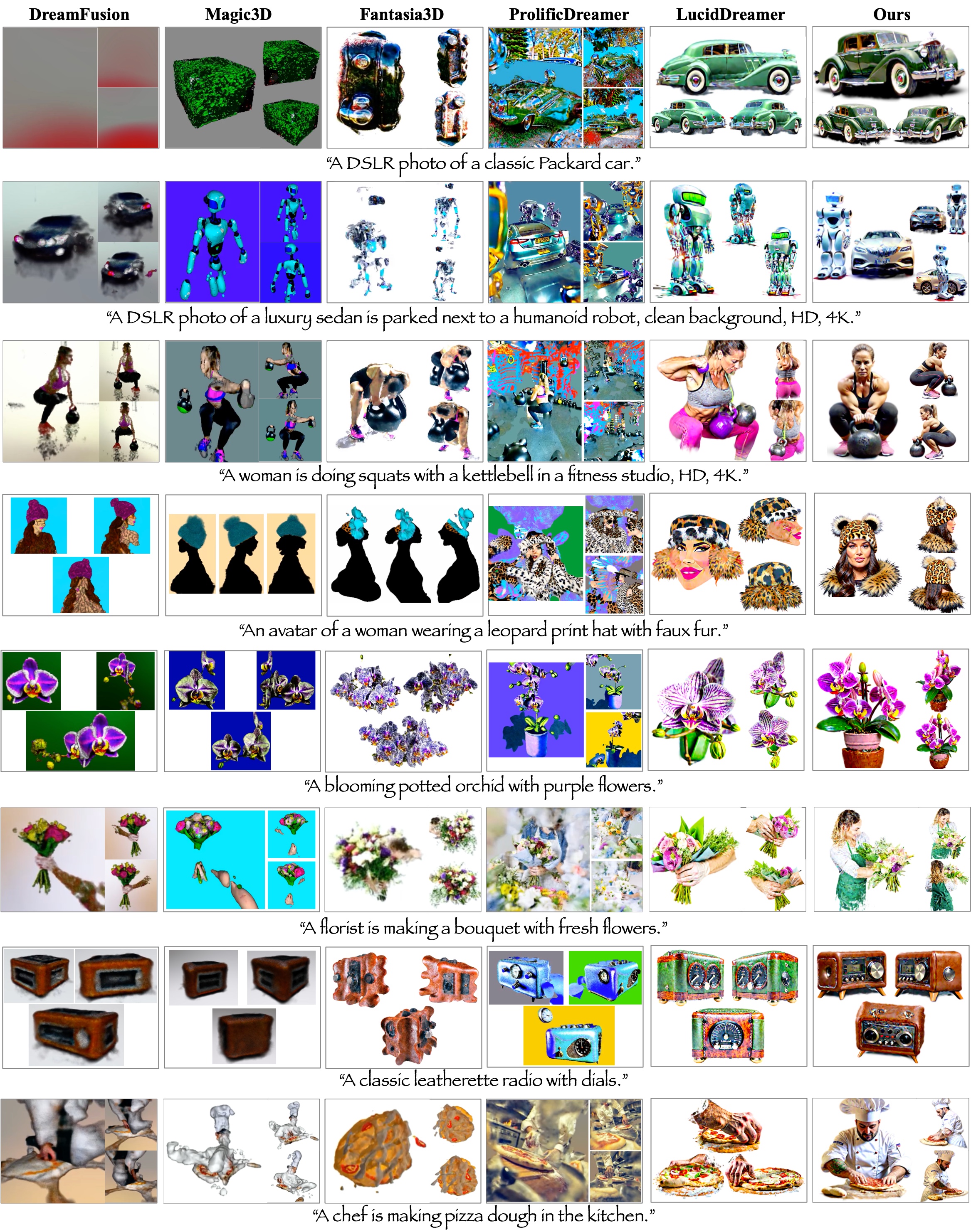}
    \caption{\textbf{Comparison of our method with existing text-to-3D baselines.} Experimental results demonstrate that our TV-3DG effectively generates complex 3D content closely aligned with the provided text prompts, characterized by high fidelity and detailed intricacy.
    }
    \label{fig:Comparison_CE} 
\end{figure*}

\subsection{Quantitative Analysis.}
\begin{figure}[t]
    \centering
    \includegraphics[width=\linewidth]{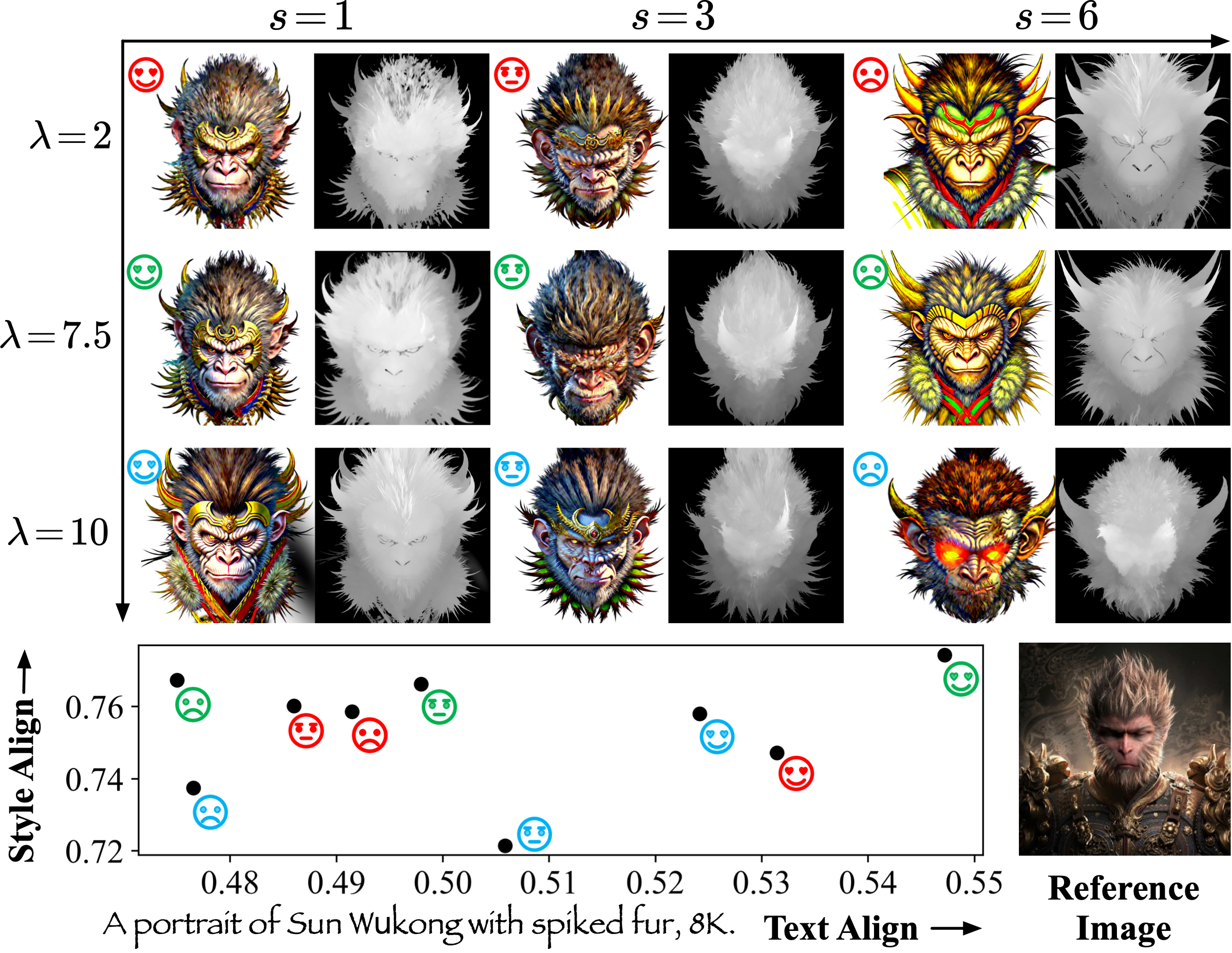}
    \caption{\textbf{Ablation on the hyperparameters} $\lambda, s$. Each scenario is marked with a corresponding emoji in the top-left corner, indicating its position in the quantitative plots of Style Align and Text Align. Here, Style Align represents the similarity of DINO \cite{caron2021emerging} features between the rendered image and the reference image, while Text Align (\ie CLIP-Score) measures the degree of alignment between the rendered image and the input text.}
    \label{fig:ablation_cfg_pag}
\end{figure}

Table~\ref{tab:clip-score-baselines} presents a detailed quantitative analysis of our framework's performance across three distinct tasks: the text-to-3D generation task (top section), the 3D stylized generation task (middle section), and an evaluation of the text-to-3D quality among various optimization algorithms within our TV-3DG framework (bottom section).
For the text-to-3D task, we source text prompts from VP3D \cite{chen2024vp3d} to ensure an objective evaluation. Our results demonstrate superior coherence and alignment with the text, with the FID metric indicating that our method's rendered images closely resemble the corresponding 2D images generated from text.
In the 3D stylized generation task, we randomly generate 20 text prompts and reference images, using the latter as visual prompts. Given the discrepancies between the reference images and the textual descriptions, we initially generate images under the dual guidance of text and reference images within style transfer frameworks \cite{hertz2024style,jeong2024visual,ye2023ip}. Subsequently, GPT-4o\footnote{Term of Service: https://openai.com/index/hello-gpt-4o} is employed to generate a prompt for the CLIP-Score evaluation.
Our method also achieves the best results in stylized generation. In Fig.~\ref{fig:ablation_cfg_pag}, the use of only original text prompts leads to CLIP-Score metrics that are relatively lower compared to those in Table~\ref{tab:clip-score-baselines}. Additionally, we employ DINO \cite{caron2021emerging} features to assess the match between the reference images and the stylized generation results.
To ensure a comprehensive assessment, each 3D object is rendered from eight equidistant viewpoints around the azimuth. The overall CLIP-Score is calculated as the mean of similarity scores between the rendered views and their corresponding text prompts.
For the user study, user preferences are assessed through rankings (lower is better) averaged over 20 samples. The results of our evaluation clearly highlight the superior performance of our TV-3DG framework, demonstrating enhanced 3D quality and better alignment with text prompts.
In Table~\ref{tab:clip-score-ablation}, we provide a detailed assessment of the metric variations corresponding to various ablation studies. A more comprehensive analysis can be found in the subsequent section dedicated to ablation experiments.

\begin{table*}[th]
\centering
\caption{Comparative analysis: 
Evaluation of CLIP-Scores across multiple CLIP retrieval models, assessment of average LPIPS across different pretrained deep networks, and results from a user study focused on text-to-3D task (top), stylize task (middle), and diverse optimization techniques (bottom).
}
\resizebox{\textwidth}{!}{
\begin{tabular}{llcccccccc}\toprule
\multicolumn{2}{c}{\multirow{2}{*}{Method}}       & \multicolumn{3}{c}{CLIP-Score}    &  & \multicolumn{2}{c}{A-LPIPS}      & \multicolumn{1}{c}{\multirow{2}{*}{FID\,$\downarrow$}} & \multicolumn{1}{c}{\multirow{2}{*}{User Study\,$\downarrow$}} \\ \cline{3-5} \cline{7-8}
\multicolumn{2}{c}{}            &\rule{0pt}{10pt} ViT-B/16\,$\uparrow$   & ViT-B/32\,$\uparrow$    & ViT-L/14\,$\uparrow$    &  & VGG\,$\downarrow$      & Alex\,$\downarrow$      & \multicolumn{1}{c}{}     & \multicolumn{1}{c}{}   \\ \midrule
Dreamfusion \cite{poole2022dreamfusion}    & \textit{$\pm$std} & 0.6514$_{\pm 0.1032}$ & 0.6732$_{\pm 0.0916}$ & 0.4964$_{\pm 0.1392}$ &  & 0.3979$_{\pm 0.1046}$ & 0.2591$_{\pm 0.0774}$ & 471.456 & 5.59$_{\pm 1.19}$ \\
Magic3d \cite{lin2023magic3d}        & \textit{$\pm$std} & 0.6524$_{\pm 0.1103}$ & 0.6351$_{\pm 0.1018}$ & 0.4686$_{\pm 0.1553}$ &  & 0.3337$_{\pm 0.0942}$ & 0.2265$_{\pm 0.0602}$ & 458.839 & 6.00$_{\pm 0.90}$ \\
Fatasia3d \cite{chen2023fantasia3d}      & \textit{$\pm$std} & 0.6447$_{\pm 0.0997}$ & 0.6333$_{\pm 0.0818}$ & 0.4991$_{\pm 0.1361}$ &  & 0.2764$_{\pm 0.1018}$ & \textbf{0.1839}$_{\pm 0.0478}$ & 459.270 & 5.19$_{\pm 1.66}$ \\
ProlificDreamer \cite{wang2024prolificdreamer} & \textit{$\pm$std} & 0.7408$_{\pm 0.0812}$ & 0.7273$_{\pm 0.0837}$ & 0.5662$_{\pm 0.0980}$ &  & 0.5015$_{\pm 0.1142}$ & 0.2692$_{\pm 0.0698}$ & 420.945 & 4.37$_{\pm 1.42}$ \\
LucidDreamer \cite{liang2023luciddreamer}   & \textit{$\pm$std} & 0.6546$_{\pm 0.0763}$ & 0.6549$_{\pm 0.0784}$ & 0.5291$_{\pm 0.0856}$ &  & 0.2845$_{\pm 0.1023}$ & 0.2702$_{\pm 0.0657}$ & 377.619 & 2.07$_{\pm 0.98}$ \\
VP3D  \cite{chen2024vp3d}          & \textit{$\pm$std} & 0.7855$_{\pm 0.0454}$ & 0.7755$_{\pm 0.0656}$ & 0.6193$_{\pm 0.0855}$ &  & 0.3996$_{\pm 0.0760}$ & 0.2757$_{\pm 0.0931}$ & 447.208 & 3.22$_{\pm 0.99}$ \\
TV-3DG (Ours)   & \textit{$\pm$std} & \textbf{0.8000}$_{\pm 0.0333}$ & \textbf{0.8089}$_{\pm 0.0342}$ & \textbf{0.6587}$_{\pm 0.0501}$ &  & \textbf{0.2558}$_{\pm 0.0823}$ & 0.2104$_{\pm 0.0761}$ & \textbf{351.511} & \textbf{1.56}$_{\pm 0.83}$ \\ \midrule
MVedit  \cite{chen2024generic}        & \textit{$\pm$std} & 0.7192$_{\pm 0.0943}$ & 0.6604$_{\pm 0.1079}$ & 0.5423$_{\pm 0.1234}$ &  & 0.2752$_{\pm 0.0447}$ & 0.1732$_{\pm 0.0445}$ & 453.222 & 2.30$_{\pm 0.66}$ \\
IPdreamer \cite{zeng2023ipdreamer}      & \textit{$\pm$std} & 0.5739$_{\pm 0.1128}$ & 0.5963$_{\pm 0.1200}$ & 0.4394$_{\pm 0.0898}$ &  & 0.3570$_{\pm 0.0759}$ & 0.2454$_{\pm 0.0775}$ & 490.225 & 2.48$_{\pm 0.69}$ \\
TV-3DG (Ours)   & \textit{$\pm$std} & \textbf{0.7792}$_{\pm 0.0866}$ & \textbf{0.7742}$_{\pm 0.1060}$ & \textbf{0.6159}$_{\pm 0.0980}$ &  & \textbf{0.2626}$_{\pm 0.0583}$ & \textbf{0.1548}$_{\pm 0.0458}$ & \textbf{444.603} & \textbf{1.22}$_{\pm 0.42}$ \\ \midrule
SDS  \cite{poole2022dreamfusion}           & \textit{$\pm$std} & 0.7675$_{\pm 0.0344}$ & 0.7898$_{\pm 0.0341}$ & 0.6570$_{\pm 0.0517}$ &  & 0.2576$_{\pm 0.0858}$ & 0.2117$_{\pm 0.0806}$ & 405.799 & 4.52$_{\pm 0.63}$ \\
CSD  \cite{yu2023csd}           & \textit{$\pm$std} & 0.7830$_{\pm 0.0315}$ & 0.7885$_{\pm 0.0381}$ & 0.6084$_{\pm 0.0569}$ &  & 0.2860$_{\pm 0.0870}$ & 0.2224$_{\pm 0.0770}$ & 353.814 & 2.70$_{\pm 1.15}$ \\
ISM \cite{liang2023luciddreamer}            & \textit{$\pm$std} & 0.7836$_{\pm 0.0336}$ & 0.7910$_{\pm 0.0380}$ & 0.6299$_{\pm 0.0490}$ &  & 0.2836$_{\pm 0.0842}$ & 0.2107$_{\pm 0.0776}$ & 381.883 & 2.85$_{\pm 1.35}$ \\
VSD \cite{wang2024prolificdreamer}            & \textit{$\pm$std} & 0.7703$_{\pm 0.0339}$ & 0.7873$_{\pm 0.0351}$ & 0.6348$_{\pm 0.0499}$ &  & 0.2718$_{\pm 0.0854}$ & 0.2159$_{\pm 0.0795}$ & 384.156 & 3.11$_{\pm 1.45}$ \\
CSM (Ours)      & \textit{$\pm$std} & \textbf{0.8000}$_{\pm 0.0333}$ & \textbf{0.8089}$_{\pm 0.0342}$ & \textbf{0.6587}$_{\pm 0.0501}$ &  & \textbf{0.2558}$_{\pm 0.0823}$ & \textbf{0.2104}$_{\pm 0.0761}$ & \textbf{351.511} & \textbf{1.81}$_{\pm 0.72}$ \\ \bottomrule
\end{tabular} }\label{tab:clip-score-baselines}
\end{table*}
\begin{figure}[t]
    \centering
    \includegraphics[width=\linewidth]{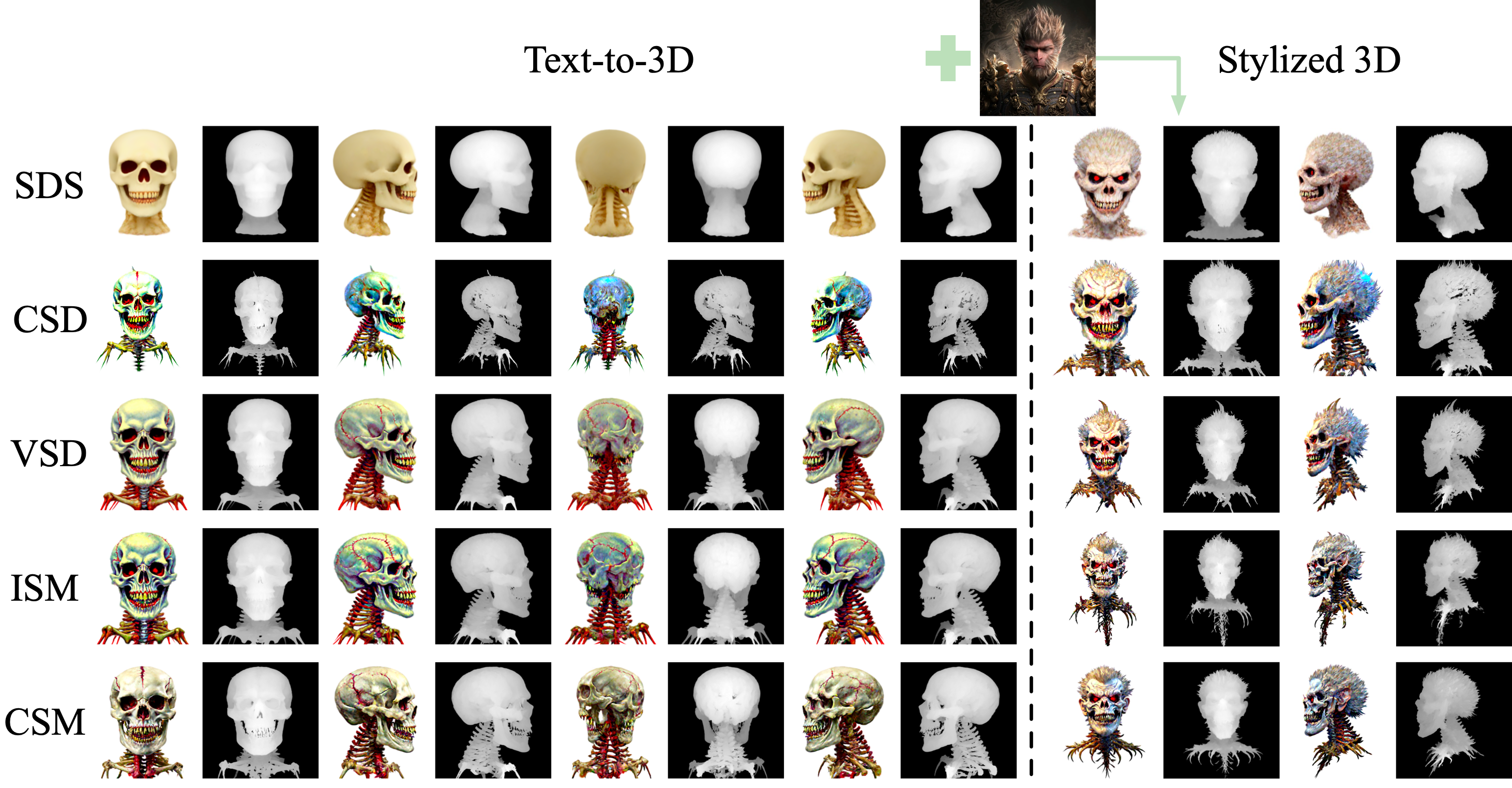}
    \caption{\textbf{Comparison with different optimization-based methods in our framework.} The text prompt is: ``A portrait of the White Bone Demon with skeletal features and a sinister grin, 8K."}
    \label{fig:vs_csm} 
\end{figure}
\begin{figure}[t]
    \centering
    \includegraphics[width=\linewidth]{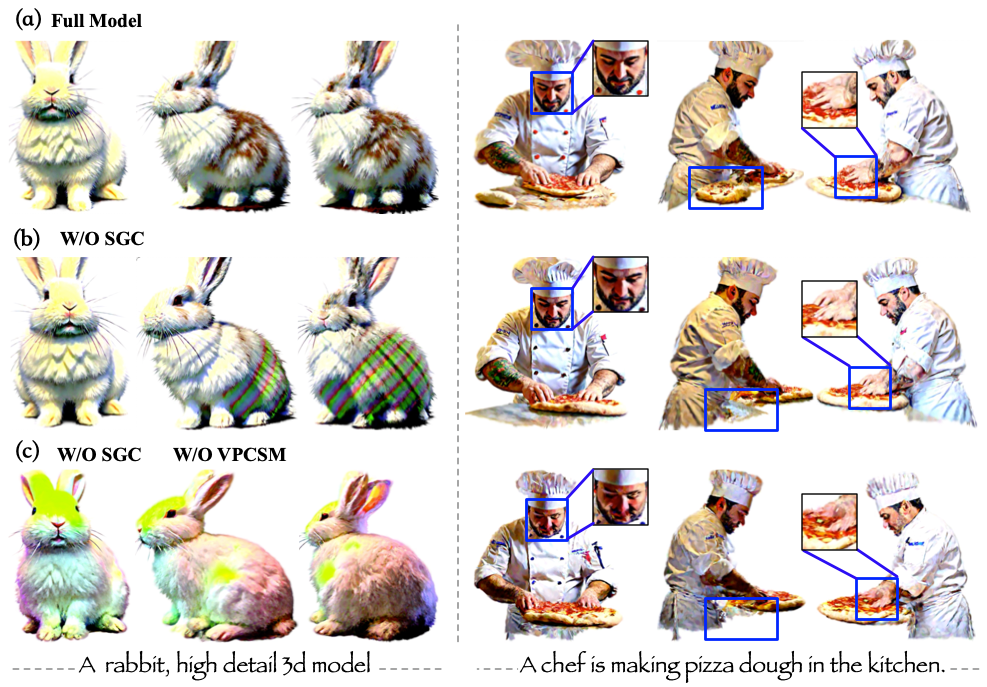}
    \caption{\textbf{Ablation on the module-wise contributions} in text-to-3D task. The absence of the SGC and VPCSM modules reduces the approach to the CSM.}
    \label{fig:ablation_ce_ip_SGC}
\end{figure}

\subsection{Qualitative Analysis}
\textbf{Visualization of TV-3DG.} Fig.~\ref{fig:visualization} shows visual results of our method across various samples with customized text and visual prompts. 
Our approach enables efficient customized generation, directly translating textual descriptions into high-quality, consistent 3D content, and also facilitating high-quality stylized 3D generation under the influence of additional visual prompts that may not semantically align with the text. 
These visualizations highlight the proficiency of our method in achieving high-quality customized 3D generation.
Some intricate reference images are sourced from Civitai \footnote{Term of Service: https://civitai.com}, with thanks to this community.
\begin{figure}[t]
    \centering
    \includegraphics[width=\linewidth]{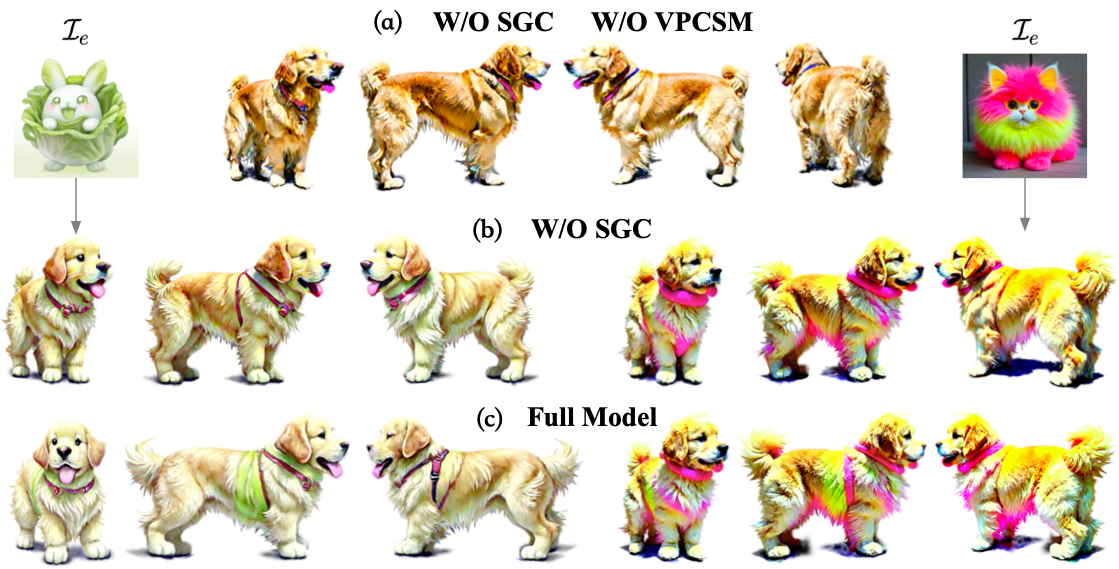}
    \caption{\textbf{Ablation on the module-wise contributions} in stylized generation. The absence of the SGC and VPCSM modules reduces the approach to the CSM algorithm.}
    \label{fig:ablation_st} 
\end{figure}

\textbf{Qualitative Comparison.}
In 3D stylized generation task, as shown in Fig.~\ref{fig:Comparison_ST}, our results exhibit superior and satisfying 3D style transfers, featuring finer texture details and more consistent geometry compared to the baseline methods.
For instance, the rabbit and golden retriever examples exhibit highly realistic stylized generation. The rabbit's body effectively learns the leopard's texture patterns while retaining its own biological features in the head. Similarly, the retriever maintains its pose and expression while adopting the fur characteristics of a cat, beyond mere texture color, showcasing remarkably realistic texture and 3D consistency. Additionally, the 3D representation of human figures is more detailed and lifelike, offering enhanced fullness and realism.
In text-to-3D task, as illustrated in Fig.~\ref{fig:Comparison_CE}, our method surpasses these baseline text-to-3D techniques by producing more plausible geometries and realistic textures. 
For instance, the Packard car example exhibits photo-realistic rendering quality with a highly authentic body texture. The color and lighting information of the Packard are harmoniously unified, whereas the Lucid Dreamer method shows issues with color oversaturation. Other comparison methods struggle to generate even the basic 3D shape.  Additionally, the example of the woman wearing a hat demonstrates exceptional hat texture details and 3D consistency.
In addition, we compare different optimization methods within our framework. As shown in Fig.~\ref{fig:vs_csm}, we demonstrate text-to-3D and stylized generation on the White Bone Demon case. SDS exhibits blurriness in both generation modes. In contrast, comparisons with other methods show that CSM achieves better customized generation.

\subsection{Ablation Studies}

\textbf{Investigation of Module-wise Contributions.}
In Fig.~\ref{fig:ablation_ce_ip_SGC} and Fig.~\ref{fig:ablation_st}, we delineate our framework into the SGC module and the VPCSM module, illustrating their respective contributions to text-to-3D tasks and stylized generation tasks.
From Fig.~\ref{fig:ablation_ce_ip_SGC}, it is evident that when the CSM algorithm fails to achieve satisfactory texture and lighting quality, the visual prompt enhances the texture, and the SGC module further generates harmonious and realistic textures and lighting conditions (\eg the rabbit example). When the CSM algorithm performs well, the remaining improvements are primarily in the realism of details (\eg the chef example).
From Fig.~\ref{fig:ablation_st}, it is clear that the visual prompt is crucial for achieving stylized generation. Additionally, while the SGC module can introduce geometric variations, its enhancement of style is limited, as the SGC primarily targets semantic and geometric improvements.

\begin{figure}[t]
    \centering
    \includegraphics[width=\linewidth]{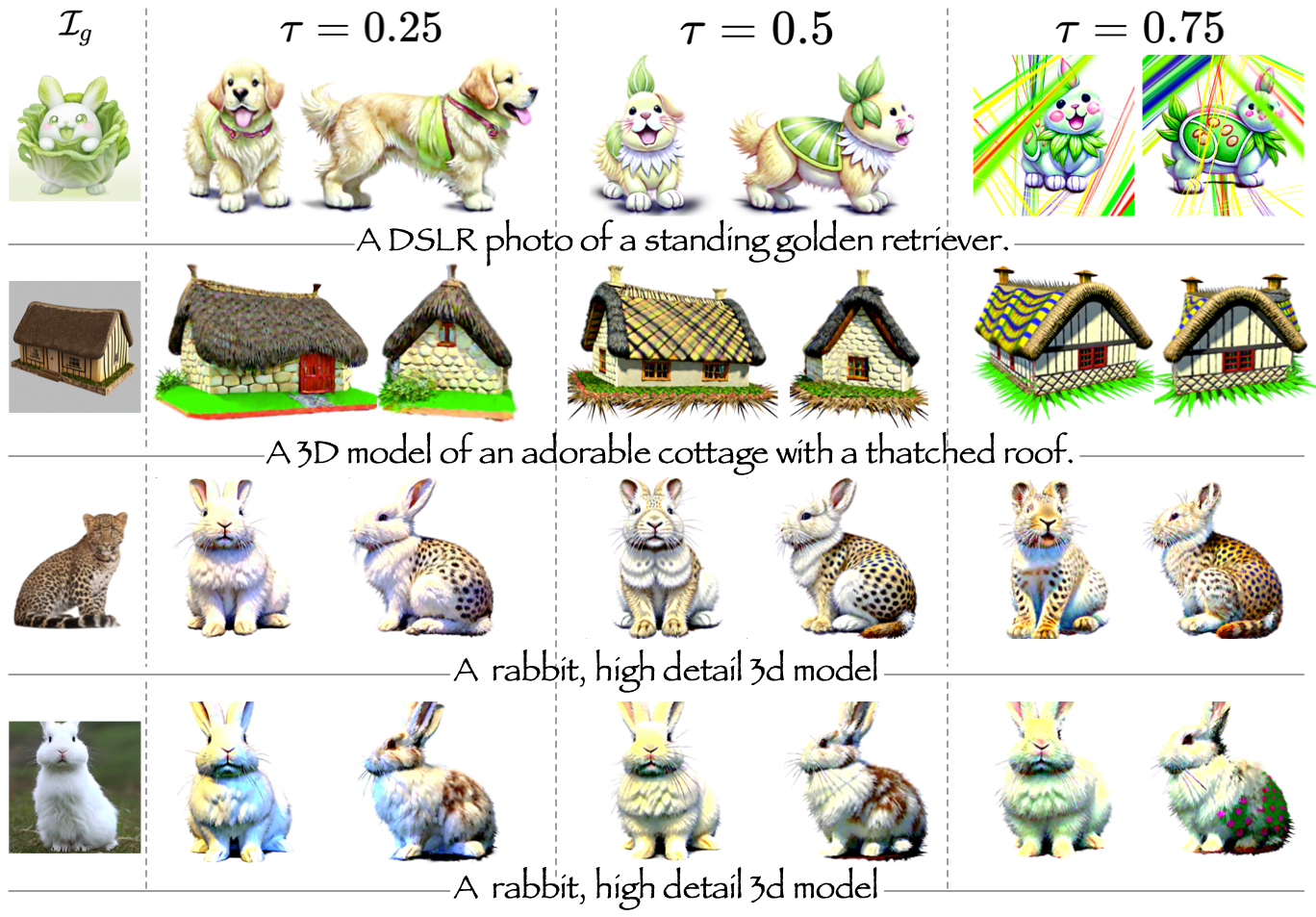}
    \caption{\textbf{Ablation on the hyperparameters} $\tau$. Setting $\tau = 0.5$ consistently reveals a trade-off in both stylized generation and text-to-3D generation.}
    \label{fig:ablation_tau}
\end{figure}

\begin{figure}[t]
    \centering
    \includegraphics[width=\linewidth]{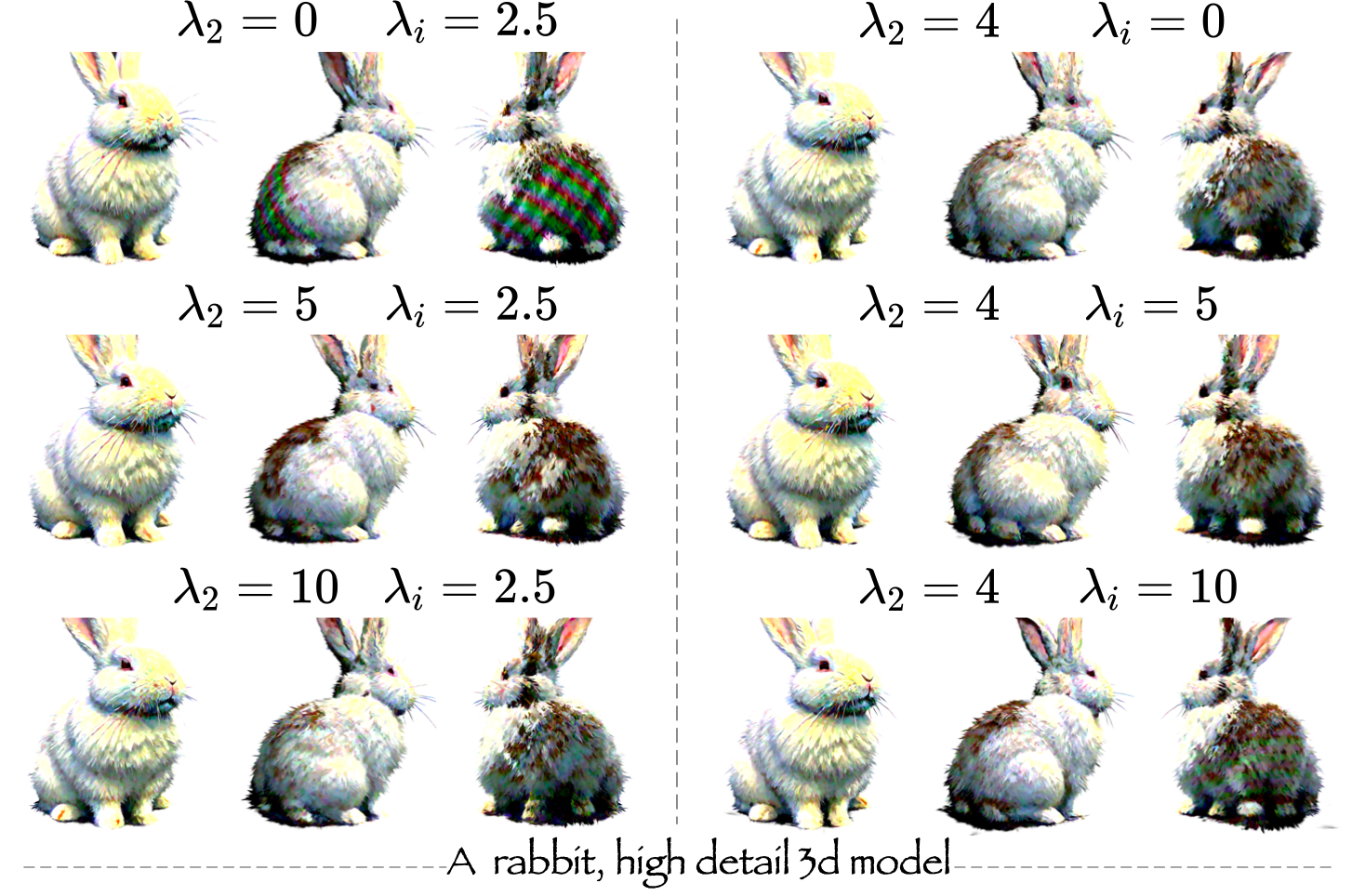}
    \caption{\textbf{Ablation on the hyperparameters} $\lambda_{2}$ and $\lambda_{i}$. A trade-off is observed when setting $\lambda_{2}=4$ and $\lambda_{i}=2.5$.}
    \label{fig:ablation_lambdas}
\end{figure}

\begin{figure}[t]
    \centering
    \includegraphics[width=\linewidth]{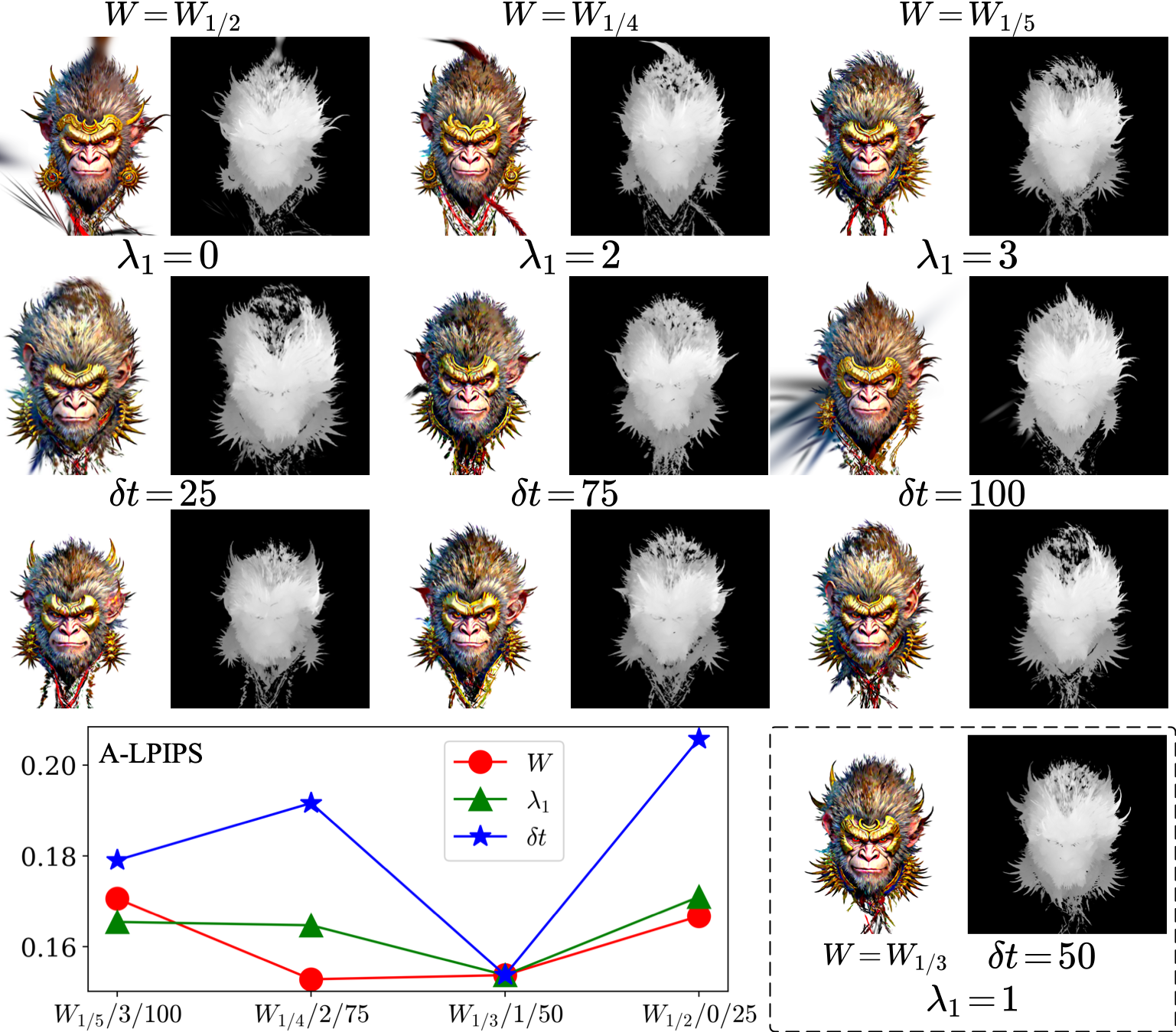}
    \caption{\textbf{Ablation on the hyperparameters} $\lambda_1, \delta t$ and $W$.}
    \label{fig:ablation_lambda1_deltat_w}
\end{figure}

\textbf{Hyperparameters.} As shown in Fig.\ref{fig:ablation_cfg_pag}, Fig.\ref{fig:ablation_tau}, Fig.\ref{fig:ablation_lambdas}, and Fig.\ref{fig:ablation_lambda1_deltat_w}, we identify a trade-off with $\tau=0.5$, $\lambda=7.5$, $\lambda_1=1$, $\lambda_{2}=4$, $\lambda_{i}=2.5$, and $s=1$.
In Fig.~\ref{fig:ablation_cfg_pag}, as $s$ increases, Wukong's portrait becomes increasingly distant from the semantic information, due to the increased perturbation leading to a deviation from the target mode. The parameter $\lambda$ is set to the standard text-to-image configuration of 7.5.
The value of $\tau$ significantly impacts both stylized generation and text-to-3D customization tasks. As shown in Fig.~\ref{fig:ablation_tau}, we find that setting $\tau$ around 0.5 achieves optimal performance in customized generation.
The values of $\lambda_2$ and $\lambda_i$ affect the realism and plausibility of 3D textures, as illustrated in Fig.~\ref{fig:ablation_lambdas}.
The choices of $\lambda_1$, $\delta t$, and $W$ influence both geometric and texture aspects, as shown in Fig.~\ref{fig:ablation_lambda1_deltat_w}. Here, $W_{1/5}$ indicates that the warmup steps are set to $\frac{1}{5}$ of the total steps. We observe that $\lambda_1=1$, $\delta t=50$, and $W=W_{1/5}$ yield more coherent textures and higher A-LPIPS scores, indicating greater consistency.

\textbf{Stylized Generation.}
In Fig.~\ref{fig:ablation_st}, we present the ablation results using two reference images applied to the same prompts as in Fig.~\ref{fig:visualization}. Without the visual prompt, the process reverts to the basic CSM algorithm for fundamental text-to-3D generation, as shown in Fig.~\ref{fig:ablation_st}(a). The combined effect of our two modules achieves the highest quality in stylized generation, offering superior texture details and consistent geometry.
As seen in Fig.\ref{fig:ablation_cfg_pag}, Fig.~\ref{fig:ablation_tau}, and Fig.~\ref{fig:ablation_lambda1_deltat_w}, the degree of stylization is primarily influenced by the hyperparameter $\tau$, while the quality of stylized content is mainly affected by other hyperparameters, such as $\delta t$ and $\lambda$.
Additionally, we conduct a quantitative experiment, as shown in Table~\ref{tab:clip-score-ablation}. Our full model achieves the highest scores across all ViT models. Conversely, the lowest scores are predominantly observed in the absence of both the SGC and VPCSM modules and in cases where the parameters significantly deviate from those of the standard full model.
In Table~\ref{tab:clip-score-ablation}, we also observe that the differences in consistency (A-LPIPS) due to varying parameters are not as pronounced as those in FID and CLIP-Score. This is because FID primarily evaluates the difference between the generated distribution and the pseudo-real distribution of images corresponding to the text prompt. CLIP-Score assesses the discrepancy between the generated content and the text, where any content modification impacts the score. In contrast, LPIPS measures intrinsic 3D visual consistency, which remains robust to changes in appearance caused by various parameter adjustments. This robustness is because the primary influence on 3D consistency originates from the CSM algorithm.

\textbf{Text-to-3D Generation.}
Fig.~\ref{fig:ablation_ce_ip_SGC} illustrates the effects of the SGC and VPCSM modules. It is evident that the rabbit's texture and shading details, as well as the chef's hand and facial details, achieve high-quality generation through the combined effect of both modules.
From Fig.~\ref{fig:ablation_tau} and Fig.~\ref{fig:ablation_lambdas}, it is clear that when the visual prompt aligns with the text meaning, parameters such as $\lambda_i$, $\lambda_2$, and $\tau$ primarily influence the quality of the generated content.
We also conduct a quantitative experiment, as shown in Table~\ref{tab:clip-score-ablation}. Our full model achieves the highest scores, except for the setting where $\lambda_2=5$ under the ViT-L/14 model, which slightly outperforms the full model.

\subsection{Evaluation details}

\textbf{2D Experiments on CSM.}
We demonstrate the efficacy of our CSM and the series of SDS loss functions in optimizing image results at the 2D level, as shown in Fig.~\ref{fig:sds-analysis} and Fig.~\ref{fig:csm-sds}. For the implementation of these experiments, we draw inspiration from the methodology employed in threestudio \cite{threestudio2023}. In practice, we use an Adam optimizer to refine the noise space, with a learning rate set to 0.001 and an optimization iteration count of 500 steps. We use the stable diffusion v1.5 \cite{rombach2022high} as our base model. The selection of timestep is uniformly sampled from a range of $t_{min}=0.02$ to $t_{max}=0.98$ (\ie $t\sim \mathcal{U}(0.02,0.98)$).

\textbf{GPT-4o for Qualitative Evaluation.}
We selected GPT-4o, the recently released multimodal model by OpenAI, as the text extractor for quantitative evaluation of our stylized experiments.
Specifically, we first input the text prompt and visual prompt into the customized 2D model, IP-Adapter \cite{ye2023ip}, to generate a standard fused 2D image. 
Then, we use GPT-4o to extract a textual description of this image.
These textual descriptions are utilized in the CLIP-Score to evaluate the multi-view images rendered from the 3D objects.

\textbf{Details of User Study.}
We designed our user evaluation questionnaire to assess various aspects of the 3D generation models. 
In the study, we anonymously and randomly sorted the 3D videos of each case and presented them to the users.
Participants evaluated the quality, consistency, aesthetic satisfaction, and alignment between the 3D content and text description in the text-to-3D task.
Additionally, they assessed the style alignment between the 3D content and the visual prompt in the stylize task.
Participants ranked the samples from different methods based on each criterion, with lower ranks indicating better performance (\ie a rank of 1 indicates the best performance).
The final score for each method was determined by averaging the ranks across all participants and criteria.
Our participants were primarily professionals in the 3D vision field.
We received a total of 27 complete responses, with participants' ages ranging from 19 to 47 years, predominantly male.

\begin{table*}[]
\caption{Ablation Analysis: We present the CLIP-Score and A-LPIPS metrics across customized generation tasks. The top three data points are highlighted in bold for emphasis. Here, FM/condition indicates the modification of parameters in the full model to the settings specified by the condition.}
\centering
\resizebox{\textwidth}{!}{
\begin{tabular}{llllllllllllll} \toprule
\multicolumn{1}{c}{} & \multicolumn{6}{c}{Text-to-3D Task} &  & \multicolumn{6}{c}{Stylize Task} \\ \cline{2-7} \cline{9-14} 
\multicolumn{1}{c}{\multirow{-2}{*}{Ablations}} &\rule{0pt}{10pt} ViT-B/16 $\uparrow$ & ViT-B/32 $\uparrow$ & ViT-L/14 $\uparrow$ & VGG $\downarrow$ & Alex $\downarrow$ & FID $\downarrow$ &  & ViT-B/16 $\uparrow$ & ViT-B/32 $\uparrow$ & ViT-L/14 $\uparrow$ & VGG $\downarrow$ & Alex $\downarrow$ & FID $\downarrow$ \\ \midrule
Full Model (FM) & \textbf{0.8000} & \textbf{0.8089} & \textbf{0.6587} & \textbf{0.2558} & 0.2104 & \textbf{351.5113} &  & \textbf{0.7790} & \textbf{0.7740} & \textbf{0.6160} & 0.2626 & 0.1548 & \textbf{444.6026} \\
w/o SGC & 0.7879 & 0.7841 & 0.6499 & 0.2932 & 0.2224 & 364.2494 &  & 0.7615 & 0.7362 & 0.5640 & 0.2602 & 0.1887 & 463.9845 \\
w/o SGC \& VPCSM & 0.7514 & 0.7506 & 0.5915 & 0.3066 & 0.2030 & 383.8356 &  & 0.5611 & 0.5544 & 0.4097 & 0.3141 & 0.1611 & 533.6832 \\ \midrule
FM/$\tau=0.25$ & 0.7987 & 0.7943 & 0.6417 & 0.2663 & 0.2604 & 361.8608 &  & 0.7048 & 0.6899 & 0.5490 & 0.2601 & 0.1823 & 457.2324 \\
FM/$\tau=0.75$ & 0.7793 & 0.7577 & \textbf{0.6568} & 0.2597 & 0.2028 & 359.5635 &  & 0.7774 & 0.7533 & 0.5897 & 0.2893 & 0.1835 & 456.2158 \\
FM/$\tau=1.0$ & 0.7473 & 0.7614 & 0.6184 & 0.2718 & 0.3017 & 398.2681 &  & \textbf{0.7907} & \textbf{0.7617} & 0.6019 & \textbf{0.1914} & \textbf{0.0893} & 495.4952 \\ \midrule
FM/$\lambda_1=0$ & 0.7282 & 0.7150 & 0.5013 & 0.2731 & 0.2367 & 368.7211 &  & 0.6852 & 0.6834 & 0.4992 & 0.2374 & 0.1874 & 459.5195 \\
FM/$\lambda_1=2$ & 0.7668 & 0.7755 & 0.5967 & 0.3148 & 0.2768 & 368.5447 &  & 0.7541 & 0.7463 & 0.6030 & 0.2013 & 0.1639 & 452.3997 \\
FM/$\lambda_1=3$ & 0.7743 & 0.7682 & 0.5849 & 0.3389 & 0.3182 & 383.4861 &  & 0.6685 & 0.6744 & 0.4867 & \textbf{0.1937} & 0.1399 & 509.6952 \\ \midrule
FM/$\lambda_2=0$ & 0.7963 & 0.7910 & 0.6526 & \textbf{0.2522} & 0.2270 & 368.3691 &  & 0.7122 & 0.7265 & 0.6021 & 0.2845 & 0.1480 & 455.5147 \\
FM/$\lambda_2=5$ & 0.7937 & \textbf{0.8087} & \textbf{0.6685} & 0.2693 & 0.2300 & 378.7211 &  & 0.6958 & 0.7386 & \textbf{0.6060} & \textbf{0.2005} & 0.1398 & 456.8429 \\
FM/$\lambda_2=10$ & 0.7907 & \textbf{0.8076} & 0.6488 & 0.3342 & 0.2944 & 388.5447 &  & 0.7096 & 0.7161 & 0.5901 & 0.2510 & 0.1600 & 471.6828 \\ \midrule
FM/$\lambda_i=0$ & 0.7985 & 0.7976 & 0.6460 & 0.2713 & 0.2060 & 380.6486 &  & 0.7516 & 0.7309 & 0.5349 & 0.2430 & 0.1253 & 478.4671 \\
FM/$\lambda_i=5$ & 0.7951 & 0.8070 & 0.6530 & 0.2811 & 0.1960 & 370.8862 &  & 0.7540 & 0.7391 & 0.5586 & 0.2645 & 0.1727 & 465.2680 \\
FM/$\lambda_i=10$ & 0.7971 & 0.8007 & 0.6250 & 0.3216 & 0.3095 & 360.1498 &  & 0.7523 & 0.7282 & 0.5740 & 0.2886 & 0.2133 & 487.0088 \\ \midrule
FM/$s=0$ & 0.7743 & 0.7682 & 0.5849 & 0.3066 & 0.2330 & 353.8356 &  & \textbf{0.8023} & \textbf{0.7590} & \textbf{0.6455} & 0.3010 & \textbf{0.1046} & \textbf{448.0088} \\
FM/$s=3$ & 0.7666 & 0.7755 & 0.5967 & 0.2989 & 0.2582 & 386.8211 &  & 0.7232 & 0.7075 & 0.5347 & 0.2938 & 0.2267 & 512.3321 \\
FM/$s=6$ & 0.7530 & 0.7462 & 0.5666 & 0.2948 & 0.2568 & 406.1704 &  & 0.6790 & 0.7001 & 0.5171 & 0.3667 & 0.2192 & 573.6203 \\ \midrule
FM/$\lambda=2$ & 0.7988 & 0.8045 & 0.6375 & 0.2616 & \textbf{0.1995} & 360.1498 &  & 0.7682 & 0.7368 & 0.5497 & 0.2558 & 0.1636 & 463.7461 \\ 
FM/$\lambda=6$ & 0.7998 & 0.7973 & 0.6554 & 0.2693 & \textbf{0.2100} & 366.1704 &  & 0.7803 & 0.7440 & 0.5892 & 0.2639 & 0.1744 & 466.6163 \\
FM/$\lambda=10$ & \textbf{0.8000} & 0.7921 & 0.6335 & 0.2598 & \textbf{0.2044} & 376.8211 &  & 0.7543 & 0.7171 & 0.5775 & 0.3140 & 0.1266 & 479.7806 \\ \midrule
FM/$\delta t=25$ & 0.7949 & 0.7963 & 0.6341 & 0.2665 & 0.2319 & 365.9247 &  & 0.7642 & 0.7297 & 0.5433 & 0.2808 & 0.1718 & 451.5886 \\ 
FM/$\delta t=75$ & \textbf{0.8020} & 0.8032 & 0.6436 & 0.3249 & 0.2669 & 371.1816 &  & 0.7644 & 0.7102 & 0.5503 & 0.3007 & 0.1708 & 459.3040 \\
FM/$\delta t=100$ & 0.7925 & 0.8043 & 0.6239 & \textbf{0.2593} & 0.2155 & 359.8915 &  & 0.7698 & 0.7394 & 0.6022 & 0.2292 & 0.1243 & 475.5582 \\ \midrule
FM/$W=W_{1/5}$ & {\color[HTML]{060607} \textbf{0.8001}} & 0.7970 & 0.6412 & 0.2996 & 0.2485 & \textbf{355.0134} &  & 0.7621 & 0.7175 & 0.5831 & 0.2715 & 0.1682 & \textbf{447.5869} \\
FM/$W=W_{1/4}$ & 0.7955 & 0.7829 & 0.6468 & 0.3112 & 0.2505 & 378.7862 &  & 0.7642 & 0.7372 & 0.5927 & 0.2894 & \textbf{0.1021} & 458.1540 \\
FM/$W=W_{1/2}$ & 0.7882 & \textbf{0.8113} & 0.6552 & 0.2712 & 0.2293 & \textbf{352.3050} &  & 0.7526 & 0.7264 & 0.5675 & 0.2843 & 0.1589 & 457.2533 \\ \bottomrule
\end{tabular}}\label{tab:clip-score-ablation}
\end{table*}

\section{Discussions}

\subsection{Applications}

\begin{figure*}[tpbh]
    \centering
    \includegraphics[width=\linewidth]{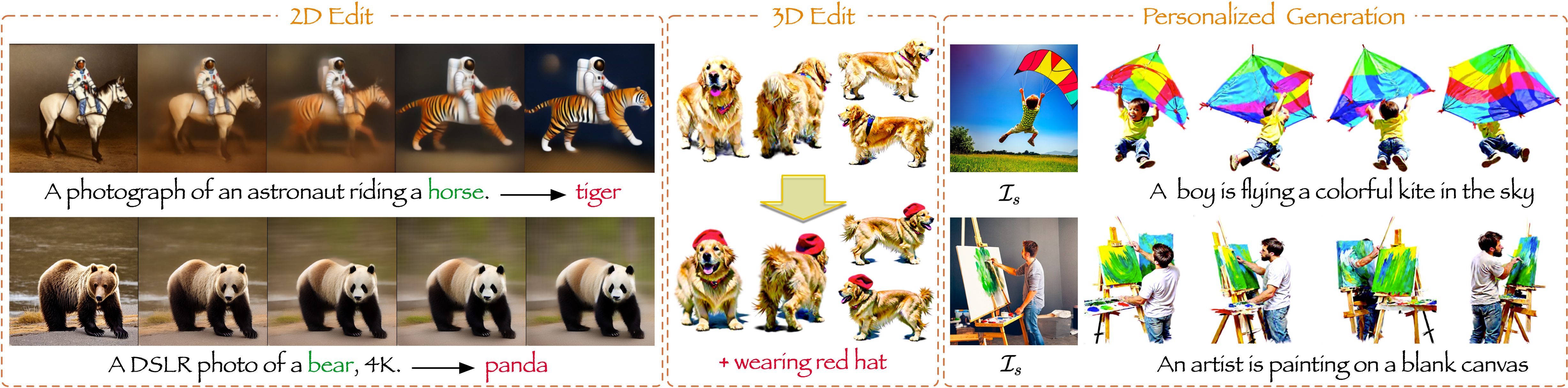}
    \caption{\textbf{Applications of our TV-3DG framework.} Our method effectively enables text-based editing in both 2D and 3D domains, and shows potential for personalized generation with identity preservation.}
    \label{fig:applications} 
\end{figure*}

\textbf{2D Editing.}
Our algorithm can effectively perform 2D editing, as demonstrated in Fig.~\ref{fig:applications}. Using the CSM algorithm, it is possible to transform objects in the original image (e.g., the horse and the bear) into objects corresponding to the given prompts (\eg the tiger and the panda), while retaining certain information such as the same actions and compositional structure. Although our algorithm is primarily designed for text-to-3D applications, it is also feasible for 2D editing. This is because the CSM algorithm optimizes the latent space of the image, using more accurate optimization gradient directions under text control information to gradually guide the latent space towards the direction described by the text. Fig.~\ref{fig:applications} illustrates the promising potential of prompt-based 2D editing.

\textbf{Personalized Generation.} 
Experimental results indicate that our method can achieve a certain degree of personalized generation. As shown in Fig.~\ref{fig:ablation_tau}, visual prompts can come from reference images (first and third rows) or from T2I-generated images consistent with text prompts (second and fourth rows). 
In text-to-3D generation, our method demonstrates alignment with the expressions or compositions of the latent images, enabling a degree of personalized generation and enhancing the controllability of customized outputs, thus proving the potential of our approach.
Additionally, Fig.~\ref{fig:applications} illustrates the capability of personalized generation. For example, the actions, poses, as well as the color schemes of the kite-flying boy and the painter, retain strong identity information in the generated 3D results. Through enhancements in semantic and geometric aspects, our method achieves consistency with the self-guidance image generated from the text, while preserving personalized features and generating in 3D.

\textbf{3D Editing.}
From our observation on various visual prompts, TV-3DG can achieve 3D editing via different visual prompts under the same text prompt, as demonstrated in the different cases shown in Fig.~\ref{fig:visualization}.
Additionally, our framework is highly effective in achieving text-guided 3D editing. As demonstrated in the middle of Fig.~\ref{fig:applications}, we use the same prompt example of a dog as before, but add the phrase ``wearing red hat." This results in a high-quality 3D editing outcome of a golden retriever that is very consistent with the original text-to-3D generation, retaining the same expression, posture, and ambiance, such as environmental lighting. The red hat is well-placed on the dog's head, showcasing strong capabilities in terms of positional accuracy, 3D consistency, and aesthetics.

\subsection{Challenges and Prospects}
While our method can generate high-quality results for most prompts, it does not universally produce corresponding high-quality outcomes for all prompts. This limitation partly arises from the core optimization-based algorithm, which distills 3D capabilities from pretrained 2D diffusion models. The effectiveness of these pretrained models is inherently limited. Future research could explore leveraging more powerful pretrained models to enhance distillation capabilities. Additionally, the SDS series algorithms optimize 3D parameters for images from each viewpoint, leading to potential inconsistencies across different perspectives. Addressing these challenges will be crucial for further advancements in achieving consistent high-quality 3D generation across a wider range of prompts.

\section{Conclusions}
In this study, we present TV-3DG, a novel text-to-3D framework that employs 2D visual prompts to generate customized, high-fidelity 3D content. Initially, we conducted an in-depth analysis of the SDS from a novel perspective and proposed an enhanced CSM algorithm, which surpasses previous SDS improvements in the domain of text-to-3D generation. Building upon the CSM, we leveraged visual prompts for controlled customized generation, integrating an attention mechanism with CFG and PAG sampling guidance techniques. We introduced the VPCSM loss to optimize the customized generation of 3D Gaussians. Furthermore, we developed the SGC module to enhance geometric and semantic outcomes in customized generation, forming the comprehensive TV-3DG system. Extensive experimental results demonstrate that our TV-3DG framework achieves high-quality customized generation, particularly in text-to-3D and stylized generation.

\bibliographystyle{IEEEtran}
\bibliography{IEEEabrv,references}

\begin{thebibliography}{10}
\providecommand{\url}[1]{#1}
\csname url@samestyle\endcsname
\providecommand{\newblock}{\relax}
\providecommand{\bibinfo}[2]{#2}
\providecommand{\BIBentrySTDinterwordspacing}{\spaceskip=0pt\relax}
\providecommand{\BIBentryALTinterwordstretchfactor}{4}
\providecommand{\BIBentryALTinterwordspacing}{\spaceskip=\fontdimen2\font plus
\BIBentryALTinterwordstretchfactor\fontdimen3\font minus \fontdimen4\font\relax}
\providecommand{\BIBforeignlanguage}[2]{{%
\expandafter\ifx\csname l@#1\endcsname\relax
\typeout{** WARNING: IEEEtran.bst: No hyphenation pattern has been}%
\typeout{** loaded for the language `#1'. Using the pattern for}%
\typeout{** the default language instead.}%
\else
\language=\csname l@#1\endcsname
\fi
#2}}
\providecommand{\BIBdecl}{\relax}
\BIBdecl

\bibitem{zhang2023adding}
L.~Zhang, A.~Rao, and M.~Agrawala, ``Adding conditional control to text-to-image diffusion models,'' in \emph{Int. Conf. Comput. Vis.}, 2023, pp. 3836--3847.

\bibitem{dhariwal2021diffusion}
P.~Dhariwal and A.~Nichol, ``Diffusion models beat gans on image synthesis,'' in \emph{Adv. Neural Inform. Process. Syst.}, vol.~34, 2021, pp. 8780--8794.

\bibitem{ho2022classifier}
J.~Ho and T.~Salimans, ``Classifier-free diffusion guidance,'' in \emph{Adv. Neural Inform. Process. Syst.}, 2022.

\bibitem{ye2023ip}
H.~Ye, J.~Zhang, S.~Liu, X.~Han, and W.~Yang, ``Ip-adapter: Text compatible image prompt adapter for text-to-image diffusion models,'' \emph{arXiv preprint arXiv:2308.06721}, 2023.

\bibitem{rombach2022high}
R.~Rombach, A.~Blattmann, D.~Lorenz, P.~Esser, and B.~Ommer, ``High-resolution image synthesis with latent diffusion models,'' in \emph{IEEE Conf. Comput. Vis. Pattern Recog.}, 2022, pp. 10\,684--10\,695.

\bibitem{ruiz2023dreambooth}
N.~Ruiz, Y.~Li, V.~Jampani, Y.~Pritch, M.~Rubinstein, and K.~Aberman, ``Dreambooth: Fine tuning text-to-image diffusion models for subject-driven generation,'' in \emph{IEEE Conf. Comput. Vis. Pattern Recog.}, 2023, pp. 22\,500--22\,510.

\bibitem{chang2015shapenet}
A.~X. Chang, T.~Funkhouser, L.~Guibas, P.~Hanrahan, Q.~Huang, Z.~Li, S.~Savarese, M.~Savva, S.~Song, H.~Su \emph{et~al.}, ``Shapenet: An information-rich 3d model repository,'' \emph{arXiv preprint arXiv:1512.03012}, 2015.

\bibitem{uy2019revisiting}
M.~A. Uy, Q.-H. Pham, B.-S. Hua, T.~Nguyen, and S.-K. Yeung, ``Revisiting point cloud classification: A new benchmark dataset and classification model on real-world data,'' in \emph{Int. Conf. Comput. Vis.}, 2019, pp. 1588--1597.

\bibitem{deitke2023objaverse}
M.~Deitke, D.~Schwenk, J.~Salvador, L.~Weihs, O.~Michel, E.~VanderBilt, L.~Schmidt, K.~Ehsani, A.~Kembhavi, and A.~Farhadi, ``Objaverse: A universe of annotated 3d objects,'' in \emph{IEEE Conf. Comput. Vis. Pattern Recog.}, 2023, pp. 13\,142--13\,153.

\bibitem{mildenhall2021nerf}
B.~Mildenhall, P.~P. Srinivasan, M.~Tancik, J.~T. Barron, R.~Ramamoorthi, and R.~Ng, ``Nerf: Representing scenes as neural radiance fields for view synthesis,'' \emph{Communications of the ACM}, vol.~65, no.~1, pp. 99--106, 2021.

\bibitem{kerbl20233d}
B.~Kerbl, G.~Kopanas, T.~Leimk{\"u}hler, and G.~Drettakis, ``3d gaussian splatting for real-time radiance field rendering,'' \emph{ACM Transactions on Graphics}, vol.~42, no.~4, pp. 1--14, 2023.

\bibitem{ho2020denoising}
J.~Ho, A.~Jain, and P.~Abbeel, ``Denoising diffusion probabilistic models,'' in \emph{Adv. Neural Inform. Process. Syst.}, vol.~33, 2020, pp. 6840--6851.

\bibitem{song2020denoising}
J.~Song, C.~Meng, and S.~Ermon, ``Denoising diffusion implicit models,'' in \emph{Int. Conf. Learn. Represent.}, 2021.

\bibitem{karras2022elucidating}
T.~Karras, M.~Aittala, T.~Aila, and S.~Laine, ``Elucidating the design space of diffusion-based generative models,'' in \emph{Adv. Neural Inform. Process. Syst.}, vol.~35, 2022, pp. 26\,565--26\,577.

\bibitem{nichol2022point}
A.~Nichol, H.~Jun, P.~Dhariwal, P.~Mishkin, and M.~Chen, ``Point-e: A system for generating 3d point clouds from complex prompts,'' \emph{arXiv preprint arXiv:2212.08751}, 2022.

\bibitem{poole2022dreamfusion}
B.~Poole, A.~Jain, J.~T. Barron, and B.~Mildenhall, ``Dreamfusion: Text-to-3d using 2d diffusion,'' in \emph{Int. Conf. Learn. Represent.}, 2022.

\bibitem{chen2023text}
Z.~Chen, F.~Wang, and H.~Liu, ``Text-to-3d using gaussian splatting,'' in \emph{IEEE Conf. Comput. Vis. Pattern Recog.}, 2024.

\bibitem{lin2023magic3d}
C.-H. Lin, J.~Gao, L.~Tang, T.~Takikawa, X.~Zeng, X.~Huang, K.~Kreis, S.~Fidler, M.-Y. Liu, and T.-Y. Lin, ``Magic3d: High-resolution text-to-3d content creation,'' in \emph{IEEE Conf. Comput. Vis. Pattern Recog.}, 2023, pp. 300--309.

\bibitem{chen2023fantasia3d}
R.~Chen, Y.~Chen, N.~Jiao, and K.~Jia, ``Fantasia3d: Disentangling geometry and appearance for high-quality text-to-3d content creation,'' in \emph{IEEE Conf. Comput. Vis. Pattern Recog.}, 2023, pp. 22\,246--22\,256.

\bibitem{tang2024lgm}
J.~Tang, Z.~Chen, X.~Chen, T.~Wang, G.~Zeng, and Z.~Liu, ``Lgm: Large multi-view gaussian model for high-resolution 3d content creation,'' in \emph{Eur. Conf. Comput. Vis.}, 2024.

\bibitem{wang2024prolificdreamer}
Z.~Wang, C.~Lu, Y.~Wang, F.~Bao, C.~Li, H.~Su, and J.~Zhu, ``Prolificdreamer: High-fidelity and diverse text-to-3d generation with variational score distillation,'' in \emph{Adv. Neural Inform. Process. Syst.}, vol.~36, 2024.

\bibitem{liang2023luciddreamer}
Y.~Liang, X.~Yang, J.~Lin, H.~Li, X.~Xu, and Y.~Chen, ``Luciddreamer: Towards high-fidelity text-to-3d generation via interval score matching,'' in \emph{IEEE Conf. Comput. Vis. Pattern Recog.}, 2024.

\bibitem{Tang_make_it_3d}
J.~Tang, T.~Wang, B.~Zhang, T.~Zhang, R.~Yi, L.~Ma, and D.~Chen, ``Make-it-3d: High-fidelity 3d creation from a single image with diffusion prior,'' in \emph{Proceedings of the IEEE/CVF International Conference on Computer Vision (ICCV)}, October 2023, pp. 22\,819--22\,829.

\bibitem{tang2023dreamgaussian}
J.~Tang, J.~Ren, H.~Zhou, Z.~Liu, and G.~Zeng, ``Dreamgaussian: Generative gaussian splatting for efficient 3d content creation,'' in \emph{Int. Conf. Learn. Represent.}, 2024.

\bibitem{yu2023csd}
X.~Yu, Y.-C. Guo, Y.~Li, D.~Liang, S.-H. Zhang, and X.~Qi, ``Text-to-3d with classifier score distillation,'' in \emph{Int. Conf. Learn. Represent.}, 2023.

\bibitem{sun2023dreamcraft3d}
J.~Sun, B.~Zhang, R.~Shao, L.~Wang, W.~Liu, Z.~Xie, and Y.~Liu, ``Dreamcraft3d: Hierarchical 3d generation with bootstrapped diffusion prior,'' in \emph{Int. Conf. Learn. Represent.}, 2023.

\bibitem{hong2023lrm}
Y.~Hong, K.~Zhang, J.~Gu, S.~Bi, Y.~Zhou, D.~Liu, F.~Liu, K.~Sunkavalli, T.~Bui, and H.~Tan, ``Lrm: Large reconstruction model for single image to 3d,'' in \emph{Int. Conf. Learn. Represent.}, 2023.

\bibitem{xu2024grm}
Y.~Xu, Z.~Shi, W.~Yifan, H.~Chen, C.~Yang, S.~Peng, Y.~Shen, and G.~Wetzstein, ``Grm: Large gaussian reconstruction model for efficient 3d reconstruction and generation,'' \emph{arXiv preprint arXiv:2403.14621}, 2024.

\bibitem{liu2023zero1to3}
R.~Liu, R.~Wu, B.~V. Hoorick, P.~Tokmakov, S.~Zakharov, and C.~Vondrick, ``Zero-1-to-3: Zero-shot one image to 3d object,'' in \emph{Int. Conf. Comput. Vis.}, 2023.

\bibitem{zhang2024clay}
L.~Zhang, Z.~Wang, Q.~Zhang, Q.~Qiu, A.~Pang, H.~Jiang, W.~Yang, L.~Xu, and J.~Yu, ``Clay: A controllable large-scale generative model for creating high-quality 3d assets,'' \emph{ACM Transactions on Graphics (TOG)}, vol.~43, no.~4, pp. 1--20, 2024.

\bibitem{xu2024sparp}
C.~Xu, A.~Li, L.~Chen, Y.~Liu, R.~Shi, H.~Su, and M.~Liu, ``Sparp: Fast 3d object reconstruction and pose estimation from sparse views,'' \emph{18th European Conference on Computer Vision (ECCV), Milano, Italy.}, 2024.

\bibitem{raj2023dreambooth3d}
A.~Raj, S.~Kaza, B.~Poole, M.~Niemeyer, N.~Ruiz, B.~Mildenhall, S.~Zada, K.~Aberman, M.~Rubinstein, J.~Barron \emph{et~al.}, ``Dreambooth3d: Subject-driven text-to-3d generation,'' in \emph{IEEE Conf. Comput. Vis. Pattern Recog.}, 2023, pp. 2349--2359.

\bibitem{chen2023control3d}
Y.~Chen, Y.~Pan, Y.~Li, T.~Yao, and T.~Mei, ``Control3d: Towards controllable text-to-3d generation,'' in \emph{Proceedings of the 31st ACM International Conference on Multimedia}, 2023, pp. 1148--1156.

\bibitem{liu2024make}
F.~Liu, H.~Wang, W.~Chen, H.~Sun, and Y.~Duan, ``Make-your-3d: Fast and consistent subject-driven 3d content generation,'' in \emph{Eur. Conf. Comput. Vis.}, 2024.

\bibitem{chen2024vp3d}
Y.~Chen, Y.~Pan, H.~Yang, T.~Yao, and T.~Mei, ``Vp3d: Unleashing 2d visual prompt for text-to-3d generation,'' in \emph{IEEE Conf. Comput. Vis. Pattern Recog.}, 2024.

\bibitem{zeng2023ipdreamer}
B.~Zeng, S.~Li, Y.~Feng, H.~Li, S.~Gao, J.~Liu, H.~Li, X.~Tang, J.~Liu, and B.~Zhang, ``Ipdreamer: Appearance-controllable 3d object generation with image prompts,'' \emph{arXiv preprint arXiv:2310.05375}, 2023.

\bibitem{chen2024generic}
H.~Chen, R.~Shi, Y.~Liu, B.~Shen, J.~Gu, G.~Wetzstein, H.~Su, and L.~Guibas, ``Generic 3d diffusion adapter using controlled multi-view editing,'' \emph{arXiv preprint arXiv:2403.12032}, 2024.

\bibitem{wang2024themestation}
Z.~Wang, T.~Wang, G.~Hancke, Z.~Liu, and R.~W. Lau, ``Themestation: Generating theme-aware 3d assets from few exemplars,'' in \emph{ACM SIGGRAPH}, 2024.

\bibitem{ahn2024self}
D.~Ahn, H.~Cho, J.~Min, W.~Jang, J.~Kim, S.~Kim, H.~H. Park, K.~H. Jin, and S.~Kim, ``Self-rectifying diffusion sampling with perturbed-attention guidance,'' in \emph{Eur. Conf. Comput. Vis.}, 2024.

\bibitem{jun2023shap}
H.~Jun and A.~Nichol, ``Shap-e: Generating conditional 3d implicit functions,'' \emph{arXiv preprint arXiv:2305.02463}, 2023.

\bibitem{zou2023triplane}
Z.-X. Zou, Z.~Yu, Y.-C. Guo, Y.~Li, D.~Liang, Y.-P. Cao, and S.-H. Zhang, ``Triplane meets gaussian splatting: Fast and generalizable single-view 3d reconstruction with transformers,'' in \emph{IEEE Conf. Comput. Vis. Pattern Recog.}, 2023.

\bibitem{song2020score}
Y.~Song, J.~Sohl-Dickstein, D.~P. Kingma, A.~Kumar, S.~Ermon, and B.~Poole, ``Score-based generative modeling through stochastic differential equations,'' in \emph{Int. Conf. Learn. Represent.}, 2021.

\bibitem{hong2024debiasing}
S.~Hong, D.~Ahn, and S.~Kim, ``Debiasing scores and prompts of 2d diffusion for view-consistent text-to-3d generation,'' in \emph{Adv. Neural Inform. Process. Syst.}, vol.~36, 2024.

\bibitem{zhong2024dreamlcm}
Y.~Zhong, X.~Zhang, Y.~Zhao, and Y.~Wei, ``Dreamlcm: Towards high quality text-to-3d generation via latent consistency model,'' in \emph{ACM Multimedia 2024}, 2024.

\bibitem{armandpour2023re}
M.~Armandpour, H.~Zheng, A.~Sadeghian, A.~Sadeghian, and M.~Zhou, ``Re-imagine the negative prompt algorithm: Transform 2d diffusion into 3d, alleviate janus problem and beyond,'' \emph{arXiv preprint arXiv:2304.04968}, 2023.

\bibitem{shi2023mvdream}
Y.~Shi, P.~Wang, J.~Ye, M.~Long, K.~Li, and X.~Yang, ``Mvdream: Multi-view diffusion for 3d generation,'' in \emph{Int. Conf. Learn. Represent.}, 2024.

\bibitem{di2024hyper}
D.~Di, J.~Yang, C.~Luo, Z.~Xue, W.~Chen, X.~Yang, and Y.~Gao, ``Hyper-3dg: Text-to-3d gaussian generation via hypergraph,'' \emph{arXiv preprint arXiv:2403.09236}, 2024.

\bibitem{ukarapol2024gradeadreamer}
T.~Ukarapol and K.~Pruvost, ``Gradeadreamer: Enhanced text-to-3d generation using gaussian splatting and multi-view diffusion,'' \emph{arXiv preprint arXiv:2406.09850}, 2024.

\bibitem{Magic123}
G.~Qian, J.~Mai, A.~Hamdi, J.~Ren, A.~Siarohin, B.~Li, H.-Y. Lee, I.~Skorokhodov, P.~Wonka, S.~Tulyakov, and B.~Ghanem, ``Magic123: One image to high-quality 3d object generation using both 2d and 3d diffusion priors,'' in \emph{Int. Conf. Learn. Represent.}, 2024.

\bibitem{purushwalkam2024conrad}
S.~Purushwalkam and N.~Naik, ``Conrad: Image constrained radiance fields for 3d generation from a single image,'' in \emph{Adv. Neural Inform. Process. Syst.}, vol.~36, 2024.

\bibitem{hertz2024style}
A.~Hertz, A.~Voynov, S.~Fruchter, and D.~Cohen-Or, ``Style aligned image generation via shared attention,'' in \emph{Proceedings of the IEEE/CVF Conference on Computer Vision and Pattern Recognition}, 2024, pp. 4775--4785.

\bibitem{jeong2024visual}
J.~Jeong, J.~Kim, Y.~Choi, G.~Lee, and Y.~Uh, ``Visual style prompting with swapping self-attention,'' \emph{arXiv preprint arXiv:2402.12974}, 2024.

\bibitem{wang2024instantstyle}
H.~Wang, Q.~Wang, X.~Bai, Z.~Qin, and A.~Chen, ``Instantstyle: Free lunch towards style-preserving in text-to-image generation,'' \emph{arXiv preprint arXiv:2404.02733}, 2024.

\bibitem{fang2024ce3d}
S.~Fang, Y.~Wang, Y.-H. Tsai, Y.~Yang, W.~Ding, S.~Zhou, and M.-H. Yang, ``Chat-edit-3d: Interactive 3d scene editing via text prompts,'' in \emph{Eur. Conf. Comput. Vis.}, 2024.

\bibitem{liu2023stylegaussian}
K.~Liu, F.~Zhan, M.~Xu, C.~Theobalt, L.~Shao, and S.~Lu, ``Stylegaussian: Instant 3d style transfer with gaussian splatting,'' \emph{arXiv preprint arXiv:2403.07807}, 2024.

\bibitem{kompanowski2024dream}
H.~Kompanowski and B.-S. Hua, ``Dream-in-style: Text-to-3d generation using stylized score distillation,'' \emph{arXiv preprint arXiv:2406.18581}, 2024.

\bibitem{ramesh2022hierarchical}
A.~Ramesh, P.~Dhariwal, A.~Nichol, C.~Chu, and M.~Chen, ``Hierarchical text-conditional image generation with clip latents,'' \emph{arXiv preprint arXiv:2204.06125}, vol.~1, no.~2, p.~3, 2022.

\bibitem{kingma2013auto}
D.~P. Kingma, ``Auto-encoding variational bayes,'' \emph{arXiv preprint arXiv:1312.6114}, 2013.

\bibitem{wu2024consistent3d}
Z.~Wu, P.~Zhou, X.~Yi, X.~Yuan, and H.~Zhang, ``Consistent3d: Towards consistent high-fidelity text-to-3d generation with deterministic sampling prior,'' in \emph{IEEE Conf. Comput. Vis. Pattern Recog.}, 2024.

\bibitem{hoogeboom2021autoregressive}
E.~Hoogeboom, A.~A. Gritsenko, J.~Bastings, B.~Poole, R.~v.~d. Berg, and T.~Salimans, ``Autoregressive diffusion models,'' in \emph{Int. Conf. Learn. Represent.}, 2021.

\bibitem{xu2024imagereward}
J.~Xu, X.~Liu, Y.~Wu, Y.~Tong, Q.~Li, M.~Ding, J.~Tang, and Y.~Dong, ``Imagereward: Learning and evaluating human preferences for text-to-image generation,'' in \emph{Adv. Neural Inform. Process. Syst.}, vol.~36, 2024.

\bibitem{yang2022banmo}
G.~Yang, M.~Vo, N.~Neverova, D.~Ramanan, A.~Vedaldi, and H.~Joo, ``Banmo: Building animatable 3d neural models from many casual videos,'' in \emph{IEEE Conf. Comput. Vis. Pattern Recog.}, 2022, pp. 2863--2873.

\bibitem{yu2024boostdream}
Y.~Yu, S.~Zhu, H.~Qin, and H.~Li, ``Boostdream: Efficient refining for high-quality text-to-3d generation from multi-view diffusion,'' \emph{arXiv preprint arXiv:2401.16764}, 2024.

\bibitem{wang2023imagedream}
P.~Wang and Y.~Shi, ``Imagedream: Image-prompt multi-view diffusion for 3d generation,'' \emph{arXiv preprint arXiv:2312.02201}, 2023.

\bibitem{caron2021emerging}
M.~Caron, H.~Touvron, I.~Misra, H.~J{\'e}gou, J.~Mairal, P.~Bojanowski, and A.~Joulin, ``Emerging properties in self-supervised vision transformers,'' in \emph{Int. Conf. Comput. Vis.}, 2021, pp. 9650--9660.

\bibitem{eftekhar2021omnidata}
A.~Eftekhar, A.~Sax, J.~Malik, and A.~Zamir, ``Omnidata: A scalable pipeline for making multi-task mid-level vision datasets from 3d scans,'' in \emph{Proceedings of the IEEE/CVF International Conference on Computer Vision}, 2021, pp. 10\,786--10\,796.

\bibitem{hertz2022prompt}
A.~Hertz, R.~Mokady, J.~Tenenbaum, K.~Aberman, Y.~Pritch, and D.~Cohen-Or, ``Prompt-to-prompt image editing with cross attention control,'' \emph{arXiv preprint arXiv:2208.01626}, 2022.

\bibitem{radford2021learning}
A.~Radford, J.~W. Kim, C.~Hallacy, A.~Ramesh, G.~Goh, S.~Agarwal, G.~Sastry, A.~Askell, P.~Mishkin, J.~Clark \emph{et~al.}, ``Learning transferable visual models from natural language supervision,'' in \emph{Int. Conf. on Mach. Learn.}\hskip 1em plus 0.5em minus 0.4em\relax PMLR, 2021, pp. 8748--8763.

\bibitem{balaji2022ediff}
Y.~Balaji, S.~Nah, X.~Huang, A.~Vahdat, J.~Song, Q.~Zhang, K.~Kreis, M.~Aittala, T.~Aila, S.~Laine \emph{et~al.}, ``ediff-i: Text-to-image diffusion models with an ensemble of expert denoisers,'' \emph{arXiv preprint arXiv:2211.01324}, 2022.

\bibitem{tewel2023key}
Y.~Tewel, R.~Gal, G.~Chechik, and Y.~Atzmon, ``Key-locked rank one editing for text-to-image personalization,'' in \emph{ACM SIGGRAPH 2023 Conference Proceedings}, 2023, pp. 1--11.

\bibitem{jain2022zero}
A.~Jain, B.~Mildenhall, J.~T. Barron, P.~Abbeel, and B.~Poole, ``Zero-shot text-guided object generation with dream fields,'' in \emph{IEEE Conf. Comput. Vis. Pattern Recog.}, 2022, pp. 867--876.

\bibitem{dosovitskiy2020image}
A.~Dosovitskiy, L.~Beyer, A.~Kolesnikov, D.~Weissenborn, X.~Zhai, T.~Unterthiner, M.~Dehghani, M.~Minderer, G.~Heigold, S.~Gelly \emph{et~al.}, ``An image is worth 16x16 words: Transformers for image recognition at scale,'' in \emph{Int. Conf. Learn. Represent.}, 2020.

\bibitem{zhang2018lpips}
R.~Zhang, P.~Isola, A.~A. Efros, E.~Shechtman, and O.~Wang, ``The unreasonable effectiveness of deep features as a perceptual metric,'' in \emph{Proceedings of the IEEE conference on computer vision and pattern recognition}, 2018, pp. 586--595.

\bibitem{simonyan2014vgg}
K.~Simonyan and A.~Zisserman, ``Very deep convolutional networks for large-scale image recognition,'' \emph{arXiv preprint arXiv:1409.1556}, 2014.

\bibitem{krizhevsky2012alex}
A.~Krizhevsky, I.~Sutskever, and G.~E. Hinton, ``Imagenet classification with deep convolutional neural networks,'' \emph{Advances in neural information processing systems}, vol.~25, 2012.

\bibitem{heusel2017gans-fid}
M.~Heusel, H.~Ramsauer, T.~Unterthiner, B.~Nessler, and S.~Hochreiter, ``Gans trained by a two time-scale update rule converge to a local nash equilibrium,'' \emph{Advances in neural information processing systems}, vol.~30, 2017.

\bibitem{threestudio2023}
Y.-C. Guo, Y.-T. Liu, R.~Shao, C.~Laforte, V.~Voleti, G.~Luo, C.-H. Chen, Z.-X. Zou, C.~Wang, Y.-P. Cao, and S.-H. Zhang, ``threestudio: A unified framework for 3d content generation,'' \url{https://github.com/threestudio-project/threestudio}, 2023.

\end{thebibliography}

\end{document}